\documentclass{article}

\usepackage{color-edits}
\addauthor{zm}{magenta}
\usepackage{tcolorbox}

\usepackage{amsmath,amsfonts,bm}

\newcommand{\bw}{\mathbf{w}}

\newcommand{\bg}{\mathbf{g}}

\newcommand{\bu}{\mathbf{u}}

\newcommand{\tr}{\textrm{Tr}}
\newcommand{\bx}{\mathbf{x}}
\newcommand{\bQ}{\mathbf{Q}}

\newcommand{\bz}{\mathbf{z}}
\newcommand{\g}{\boldsymbol{\mathrm{g}}}

\newcommand{\e}{\mathbf{e}}

\newcommand{\ul}{\underline}

\newcommand{\ulc}{\ul{c}}
\newcommand{\barc}{\bar{c}}

\def\eqref#1{equation~(\ref{#1})}

\def\1{\bm{1}}

\DeclareMathAlphabet{\mathsfit}{\encodingdefault}{\sfdefault}{m}{sl}
\SetMathAlphabet{\mathsfit}{bold}{\encodingdefault}{\sfdefault}{bx}{n}

\newcommand{\E}{\mathbb{E}}

\newcommand{\R}{\mathbb{R}}

 \providecommand{\ttheta}{\bm{\theta}}

  \renewcommand{\gg}{\boldsymbol{g}}

  \providecommand{\e}{\mathrm{e}}

  \providecommand{\cD}{\mathcal{D}}

  \providecommand{\cL}{\mathcal{L}}
  
  \providecommand{\cN}{\mathcal{N}}

\newcommand{\lin}[1]{\ensuremath \left\langle #1 \right\rangle}

\def\remark{\addtocounter{remark}{1}\def\@currentlabel{\theremark}%
\emph{Remark~\theremark}. } \makeatother
\newcounter{remark}

 \usepackage{color-edits}
 \addauthor{mh}{red}
 \addauthor{xw}{magenta}
 \addauthor{ne}{brown}
 
\newcommand{\ols}[1]{\mskip.5\thinmuskip\overline{\mskip-.5\thinmuskip {#1} \mskip-.5\thinmuskip}\mskip.5\thinmuskip} %

\usepackage{microtype}
\usepackage{booktabs}
\usepackage{adjustbox}
\usepackage{comment}
\usepackage{caption}
\usepackage{enumitem}
\usepackage{graphicx}
\usepackage{subfigure}
\usepackage{booktabs} %
\usepackage{algorithm}
\usepackage{algorithmic}
\usepackage{wrapfig}

\usepackage{mathtools}

\usepackage{hyperref}

\definecolor{lightblue}{HTML}{B0E0E6}
\definecolor{lightorange}{HTML}{FFE4B5}
\definecolor{lightred}{HTML}{FFE4E1}
\definecolor{darkblue}{HTML}{003153}

\newcommand{\mystrut}{\rule{0pt}{0.5em}}

\newtcbox{\lightorangebox}{on line, boxrule=0pt, boxsep=0pt, top=0pt,
left=0pt, bottom=0pt, right=0pt, colback=lightorange, colframe=white,
fontupper={\mystrut}}

\newtcbox{\lightbluebox}{on line, boxrule=0pt, boxsep=0pt, top=0pt,
left=0pt, bottom=0pt, right=0pt, colback=lightblue, colframe=white,
fontupper={\mystrut}}

\newtcbox{\lightredbox}{on line, boxrule=0pt, boxsep=0pt, top=0pt,
left=0pt, bottom=0pt, right=0pt, colback=lightred, colframe=white,
fontupper={\mystrut}}

\usepackage{amsmath}
\usepackage{amssymb}
\usepackage{mathtools}
\usepackage{amsthm, enumitem}
\usepackage[capitalize,noabbrev]{cleveref}

\theoremstyle{plain}
\newtheorem{theorem}{Theorem}[section]

\newtheorem{lemma}[theorem]{Lemma}

\theoremstyle{definition}
\newtheorem{definition}[theorem]{Definition}
\newtheorem{assumption}[theorem]{Assumption}
\newtheorem{example}[theorem]{Example}
\theoremstyle{remark}
\PassOptionsToPackage{numbers, compress}{natbib}

\usepackage[preprint]{neurips_2025}

\usepackage[utf8]{inputenc} %
\usepackage[T1]{fontenc}    %
\usepackage{hyperref}       %
\usepackage{url}            %
\usepackage{booktabs}       %
\usepackage{amsfonts}       %
\usepackage{nicefrac}       %
\usepackage{microtype}      %
\usepackage{xcolor}         %

\title{\raisebox{-1.4ex}{\includegraphics[height=2em]{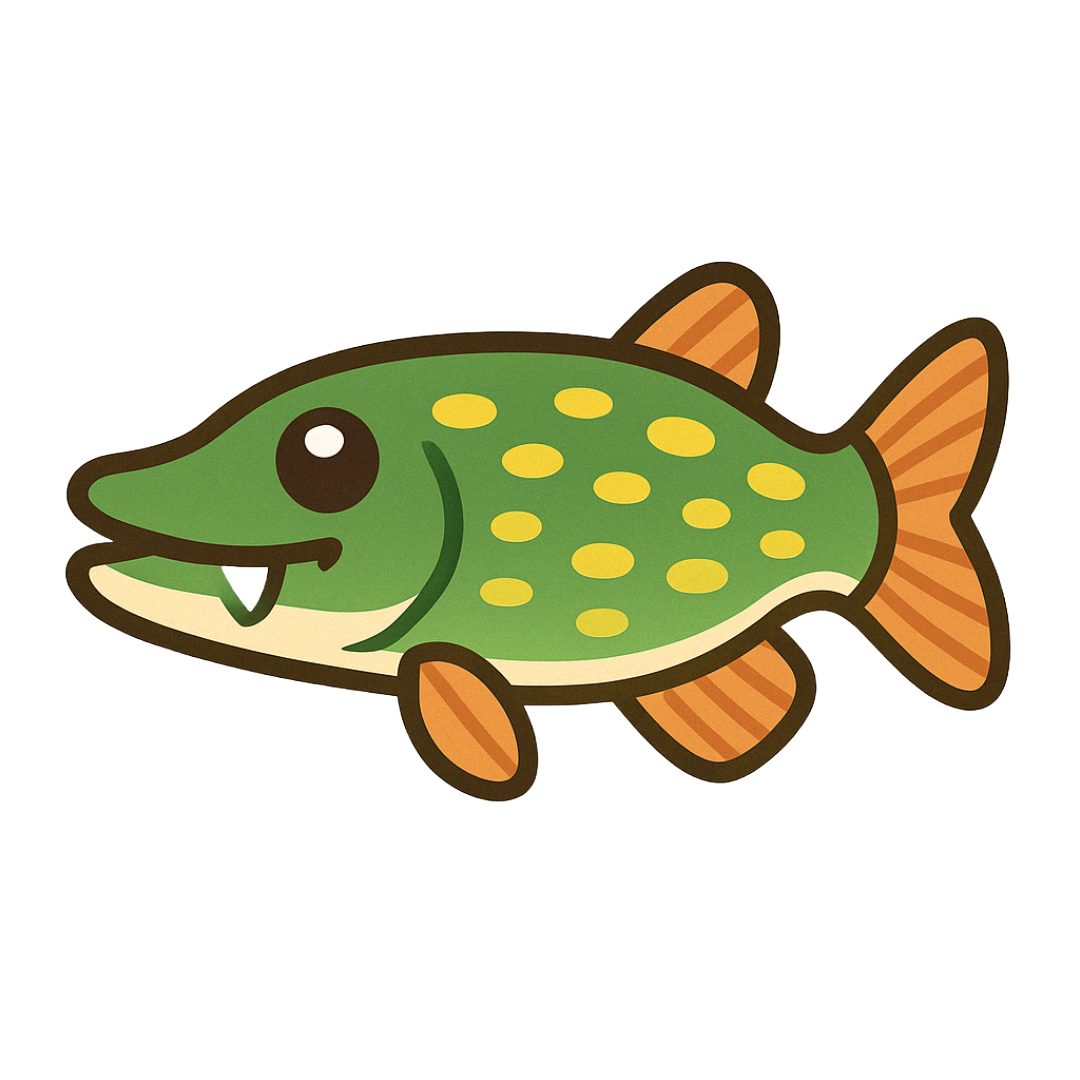}} PiKE: Adaptive Data Mixing for Large-Scale Multi-Task Learning Under Low Gradient Conflicts}

\author{
Zeman Li$^{1,2}$\thanks{Work done while interning at Google Research.} \quad Yuan Deng$^{2}$ \quad Peilin Zhong$^2$ \quad Meisam Razaviyayn$^{1,2}$ \quad Vahab Mirrokni$^2$ \\
$^1$University of Southern California \quad $^2$Google Research\\
\texttt{\{zemanli,razaviya\}@usc.edu}\\
\texttt{\{dengyuan,peilinz,mirrokni\}@google.com}
}

\begin{document}

\maketitle

\vspace{-0.6cm}

\begin{abstract}

\vspace{-0.2cm}

Modern foundation models are trained on diverse datasets to enhance generalization across tasks and domains. A central challenge in this process is determining how to effectively mix and sample data from multiple sources. This naturally leads to a multi-task learning (MTL) perspective. While prior work in MTL has emphasized mitigating gradient conflicts, we observe that large-scale pretraining scenarios—such as multilingual or multi-domain training—often exhibit little to no gradient conflict. Motivated by this observation, we propose \textbf{PiKE} (\textbf{P}ositive gradient \textbf{i}nteraction-based \textbf{K}-task weights \textbf{E}stimator), an adaptive data mixing algorithm that dynamically adjusts sampling weights during training. PiKE exploits non-conflicting gradient interactions to minimize a near-tight upper bound on the average loss decrease at each step, while incurring negligible computational overhead. We provide theoretical convergence guarantees and  show that PiKE  outperforms static and non-adaptive mixing baselines. Furthermore, we extend PiKE to promote balanced learning across tasks. Extensive experiments on large-scale language model pretraining confirm that PiKE achieves faster convergence and improved downstream performance compared to existing approaches.

\end{abstract}

\section{Introduction}

\vspace{-0.3cm}

Foundation models, such as large language models (LLMs), owe their strong generalization and multitask abilities to pretraining on diverse datasets spanning multiple domains~\citep{team2024gemini, liu2024deepseek, chowdhery2022palm}. The effectiveness of these models depend heavily on the composition of their training data~\citep{hoffmann2022empirical, du2022glam}. However, standardized practices for curating optimal pretraining data are lacking. Common approaches involve heuristic filtering, deduplication, and categorization into heterogeneous domains (e.g., The Pile~\cite{gao2020pile} has 22 domains; GLaM~\cite{du2022glam} has 6). Even after such preprocessings, determining the optimal data mixing remains a key challenge—amplified by the scale of modern models and datasets. %

A common strategy in pre-training is to use fixed data mixtures, typically chosen heuristically or via smaller proxy models. For example, mT5~\citep{xue2020mt5} weights datasets by relative size, while GLaM~\citep{du2022glam} uses downstream performance from proxy models. DoReMi~\citep{xie2024doremi} also relies on proxy models, using group distributionally robust optimization (group DRO) to set dataset weights. However, these approaches have notable limitations. First, their optimality is unclear: heuristic methods lack theoretical backing, and policies from small models may not necessarily transfer to larger ones~\citep{ye2024data}. Second, proxy models introduce substantial computational overhead, often scaling linearly or worse with the number of domains. Third, static weights fixed at initialization may become suboptimal as training progresses~\citep{zhang2024transformers, li2018measuring}, a limitation discussed further in Section~\ref{sec: building_blocks_3}.

In this work, we frame adaptive data mixture selection problem as a multitask optimization problem, enabling a principled approach to dynamically adjusting data mixture. This view is natural: each data domain typically is related to a (set of) tasks and yields a distinct gradient. Prior multitask optimization methods~\citep{yu2020gradient, wang2020gradient, chen2018gradnorm, desideri2012multiple} focus on resolving gradient conflicts that impede convergence. However, most are impractical for current LLMs  due to their $O(Kd)$ memory cost for storing $K$ gradients or $O(K^2)$ computation from repeated backpropagation. A notable exception is FAMO~\citep{liu2024famo}, which scales more efficiently. Additionally, most MTL methods are tailored for vision tasks where gradient conflicts are prevalent, a condition less common in LLM pretraining:
For example, GradVaccine~\citep{wang2020gradient} observed that in multilingual BERT (178M parameters), task gradients are mostly positively aligned or nearly orthogonal. We extend this finding to much larger autoregressive, decoder-only models (e.g., 1B parameters), which better reflect modern LLMs. Our results show minimal gradient conflict in  multilingual and multi-domain pretraining  (Section~\ref{sec: building_blocks_2}), suggesting that modern LLMs naturally exhibit cooperative (or sometimes nearly-orthogonal) gradient structures—potentially reducing the need for explicit conflict-mitigation. This motivates the central question of our work:

\vspace{-0.1cm}

\begin{center}
\emph{Can we design an effective \textbf{adaptive data mixing strategy} based on \textbf{multitask optimization}, that \textbf{exploits non-conflicting} nature of gradients and scales to \textbf{large settings} with \textbf{minimal memory and computational overhead}?}
\end{center}

\vspace{-0.1cm}

To address this, we propose \textit{PiKE}, an adaptive data mixing  that is \textbf{empirically effective} and \textbf{scalable} for pretraining LLMs. %
PiKE enjoys \textbf{theoretical guarantees} while remaining practical for large-scale settings—with many tasks, large models, and diverse data—at \textbf{negligible memory and compute overhead}. We summarize PiKE’s key features below; related work is discussed in Appendix~\ref{sec: related}.

\begin{figure*}[!tb]
\centering
\begin{minipage}{.5\textwidth}
  \centering
  \includegraphics[width=1\linewidth]{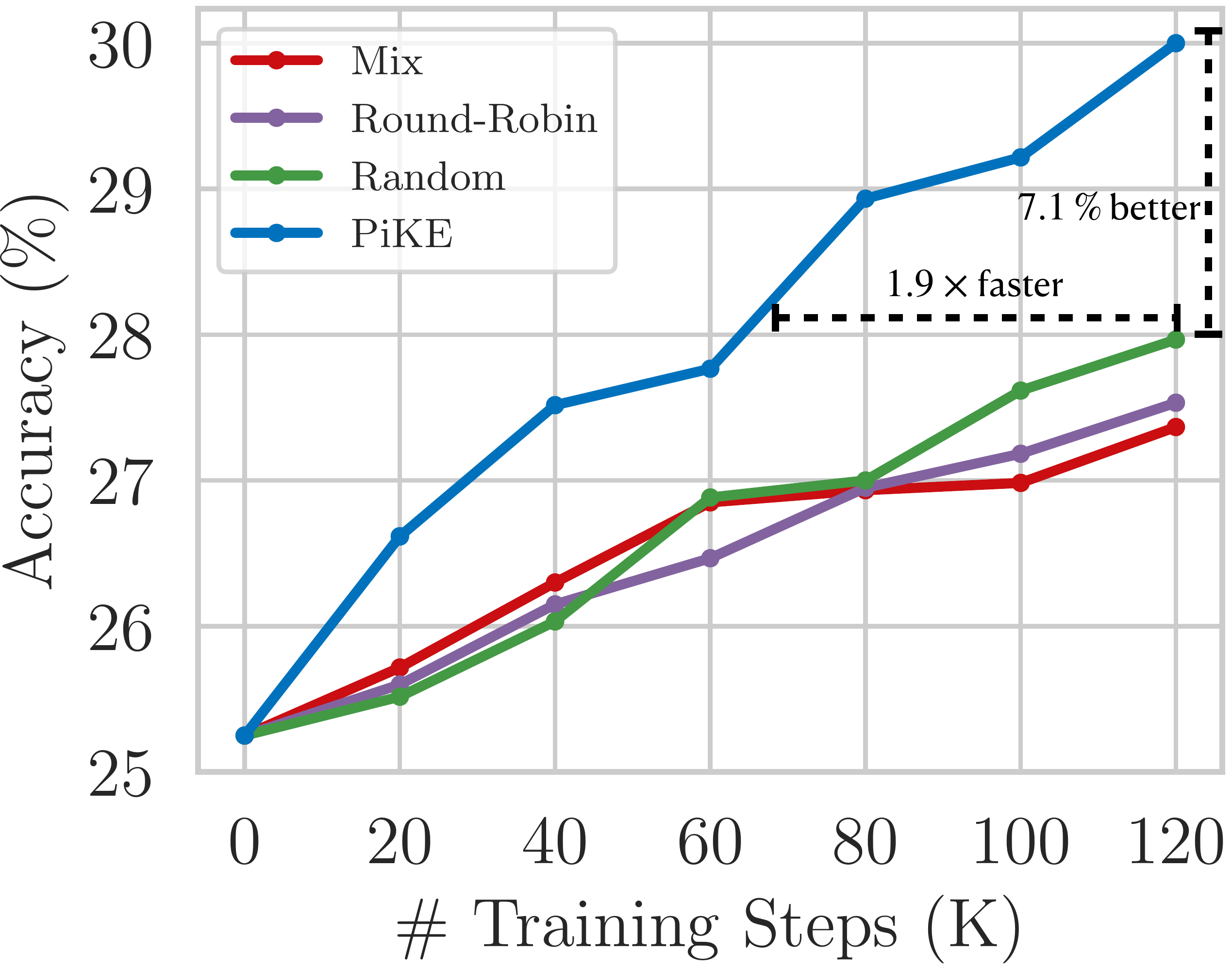}
\end{minipage}%
\begin{minipage}{.5\textwidth}
  \centering
  \includegraphics[width=1\linewidth]{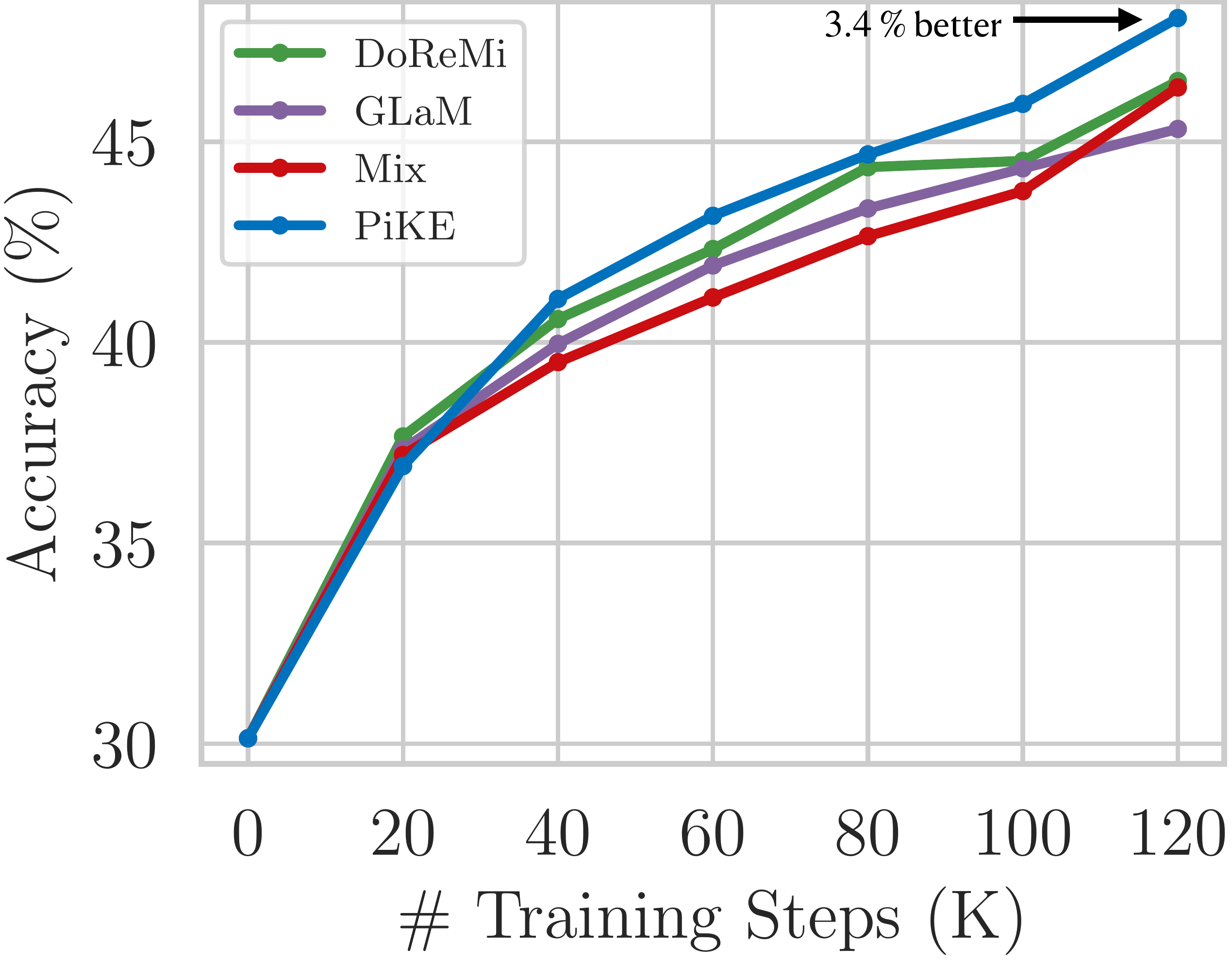}
\end{minipage}
\caption{\footnotesize 
PiKE \textbf{adaptively optimizes task weights} in pre-training, outperforming baselines. \textbf{Left (1B models, multilingual C4 en/hi):} PiKE boosts average downstream accuracy by 7.1\% and reaches baseline accuracy $1.9\times$ faster. \textbf{Right (750M models, GLaM six domains):} PiKE achieves a 3.4\% higher accuracy over DoReMi~\cite{xie2024doremi}. PiKE scales efficiently with model size and number of tasks. Detailed results are in Tables~\ref{tab:main-table-multilingual} and~\ref{tab:main-table-glam}.}
\label{fig:the-main-plots}
\end{figure*}

\vspace{-0.1cm}
\begin{tcolorbox}[colback=white,colframe=darkblue,title=\textbf{Key Features of PiKE}, titlerule=0.2mm] %
\begin{enumerate}[leftmargin=*]
    \item Adaptively adjusts mixture weights using \lightbluebox{per-task gradient magnitude} and \lightorangebox{variance}.%
    \item Theoretically, achieves near-optimal per-iteration objective decrease and enjoys  convergence guarantees (\Cref{sec:method}).
    \item Incorporates tilted empirical risk minimization~\citep{li2020tilted, mo2000fair}, promoting balanced learning across tasks and preventing task under-representation (Section~\ref{sec:fair-pike}).
    \item Scales efficiently (linearly) with model size and the number of tasks (Section~\ref{sec:exp}).
    \item Consistently outperforms existing methods across different scales (110M to 1B parameters) and scenarios (multilingual to domain mixing) (Figure~\ref{fig:the-main-plots} and Section~\ref{sec:exp}). 
\end{enumerate}
\end{tcolorbox}

\normalsize

\vspace{-0.2cm}
\paragraph{Problem Definition and Notations} \label{sec:Notations}
We aim to train a \emph{single} model with parameters \(\ttheta \in \mathbb{R}^d\) to perform \(K \geq 2\) tasks simultaneously. Each task~$k$ is associated with a smooth (possibly non-convex) loss  \(\ell_k(\ttheta, x): \mathbb{R}^d \times \mathbb{R}^{d_x} \to \mathbb{R}\) where $x$ is the data point. It is common to minimize  the total  loss:

\vspace{-0.2cm}

\begin{equation}  \label{eq:prob_def}
\min_{\ttheta \in \mathbb{R}^d} \cL(\ttheta) := \sum_{k=1}^{K} \mathbb{E}_{x \sim \cD_k} [\ell_k(\ttheta; x)],  
\end{equation}  

\vspace{-0.2cm}

where \(\cD_k\) represents the data distribution for task \(k\). We define \(\cL_k(\ttheta) := \mathbb{E}_{x \sim \cD_k} [\ell_k(\ttheta; x)] \).
For notation, \(\|\cdot\|\) represents the Euclidean norm, \(\tr(\cdot)\) denotes the trace operator, and a function \(h\) is \(L\)-Lipschitz if \(\|h(\ttheta) - h(\ttheta')\| \leq L \|\ttheta - \ttheta'\|\) for any \(\ttheta, \ttheta'\) in the domain of $h(\cdot)$. A function \(f(\cdot)\) is \(L\)-smooth if its gradient is \(L\)-Lipschitz continuous.

\section{Main Building Blocks for PiKE} \label{sec: main_building_blocks}
This section presents our main observations which form the main building blocks of PiKE.
\subsection*{Bulding Block \#1: Mixing Domains per Batch Improves LLM Generalization} \label{sec: building_blocks_1}

Optimizing~\eqref{eq:prob_def} with stochastic methods, such as Adam or SGD, requires forming batches from one or more tasks at each step. The batch selection strategy strongly influences model performance~\citep{bengio2009curriculum, ge2024data, ye2024data, xie2024doremi, liu2024regmix}. Even with fixed data proportions, a key question remains: \textit{how should one form a batch from $K$ data domains at each step?}

Batch construction is a critical design choice in multitask training, as it directly impacts learning dynamics and final model performance~\citep{bengio2009curriculum, ge2024data, ye2024data, xie2024doremi, liu2024regmix}. We focus on three standard strategies: \textit{Random}, \textit{Round-Robin}, and \textit{Mix}. Let us first define these methods assuming static, uniform sampling weights (i.e., $1/K$ per task). Let $\mathbf{b}_t = (b_{t,1}, \dots, b_{t,K})$ denote the number of samples from each task $\mathcal{D}_k$ at iteration $t$, with total batch size \(b = \sum_{k=1}^K b_{t,k}\). The strategies are defined as:
\[
\mathbf{b}_t =
\begin{cases}
    b \cdot \mathbf{e}_{k^*} & \text{\textit{Random}, where } k^* \sim \text{Uniform}(\{1, \ldots, K\}) \\
    b \cdot \mathbf{e}_{(t \bmod K) + 1} & \text{\textit{Round-Robin}} \\
    b \cdot \left( \frac{1}{K}, \ldots, \frac{1}{K} \right) & \text{\textit{Mix}}
\end{cases}
\]
where $\mathbf{e}_k \in \mathbb{R}^K$ is the $k$-th standard basis vector. That is, \textit{Random} selects one domain per batch, \textit{Round-Robin} cycles through domains, and \textit{Mix} includes all tasks in each batch.

\begin{figure}[!tb]
\vskip 0.2in
\begin{minipage}{.5\textwidth}
     \centering
    \includegraphics[width=1\columnwidth]{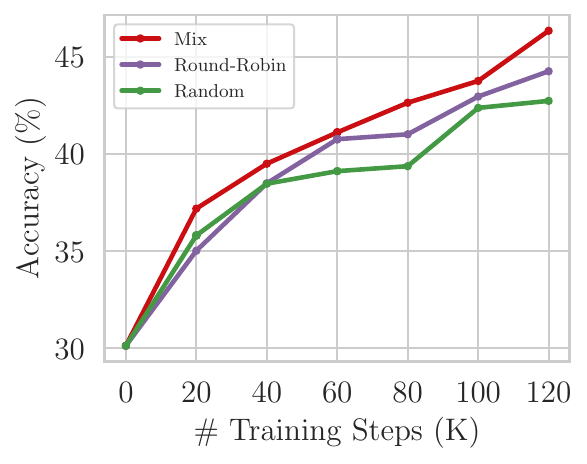}
\end{minipage}%
\begin{minipage}{.5\textwidth}
  \centering
  \includegraphics[width=1\linewidth]{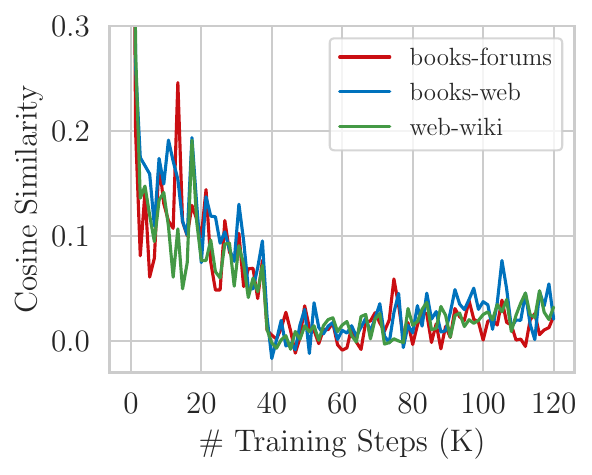}
\end{minipage}
\caption{\footnotesize \textbf{Left:} Average accuracy on four downstream tasks (ArcE, CSQA, HellaSwag, PIQA) for 750M GPT-2 style models pre-trained with \textit{Mix}, \textit{Round-Robin}, and \textit{Random} strategies. See Appendix~\ref{app: mix_rr_random} for more. \textbf{Right:} Task gradient cosine similarity for a 750M GPT-2 style model pre-trained on GLaM datasets. ``\textit{data1-data2}'' indicates gradient similarity between tasks \textit{data1} and \textit{data2}. Further results in Appendix~\ref{app: cos_sim_grad_conflict}.}
\label{fig: motivation_why_mix}
\end{figure}

Historically, \textit{Mix} has been widely used in computer vision multitask settings~\citep{dai2016instance, misra2016cross, chen2018gradnorm, ruder2019latent, yu2020gradient, liu2024famo}, while earlier language modeling efforts favored \textit{Random} or \textit{Round-Robin}~\citep{liu2015representation, luong2015multi, liu2019multi}. Recent studies on large-scale language models~\citep{devlin2018bert, raffel2020exploring, brown2020language, team2023gemini} have revisited this question and found that \textit{Mix} generally performs best—especially in diverse-data settings~\citep{du2022glam, chowdhery2023palm, xie2024doremi, gao2020pile, wang2019superglue}.
Our experiments reaffirm this trend: across a wide range of tasks and scales, \textit{Mix} consistently outperforms the alternatives (Figure~\ref{fig: motivation_why_mix}, Figure~\ref{fig:app_mix_round_rr_comparison}). This observation underpins the design of our proposed method, \textit{PiKE}.

\normalsize

\subsection*{Bulding Block \#2: Large-Scale Training: Low Gradient Conflict \& MTL Scalability Challenges} \label{sec: building_blocks_2}

The \textit{Mix} batching strategy offers a natural lens for analyzing data mixing in multitask learning (MTL). When a batch of total size $b$ is constructed with $b_k$ samples from each task $k$, the overall gradient at iteration $t$, denoted $\mathbf{g}_t$, is the average of the individual sample gradients:
\begin{equation} \label{eq: mix_framework}
\mathbf{g}_t = \frac{1}{b} \sum_{k=1}^{K} \sum_{i=1}^{b_k} \nabla \ell_k (\boldsymbol{\theta}_t; x_{k,i}) = \sum_{k=1}^K \frac{b_k}{b} \bar{\mathbf{g}}_{t,k},
\end{equation}
where $\bar{\mathbf{g}}_{t,k} = \frac{1}{b_k}\sum_{i=1}^{b_k} \nabla \ell_k(\boldsymbol{\theta}_t; x_{k,i})$  is the average gradient from task $k$, and $x_{k,i} \sim \mathcal{D}_k$. This formulation highlights how batch gradients blend contributions from different tasks and serves as the foundation for analyzing task interactions.

A central challenge in prior MTL literature is  \textit{gradient conflict}~\citep{yu2020gradient, wang2020gradient, navon2022multi}, where task gradients oppose the overall update direction. Formally, a conflict exists if:
$
\langle \mathbf{g}_t, \bar{\mathbf{g}}_{t,k} \rangle < 0,
$
indicating that the shared update could increase the loss for task $k$.

\begin{wrapfigure}{r}{0.47\textwidth}
\vspace{-0.4cm}
 \begin{center}  
\centerline{\includegraphics[width=0.5\columnwidth]{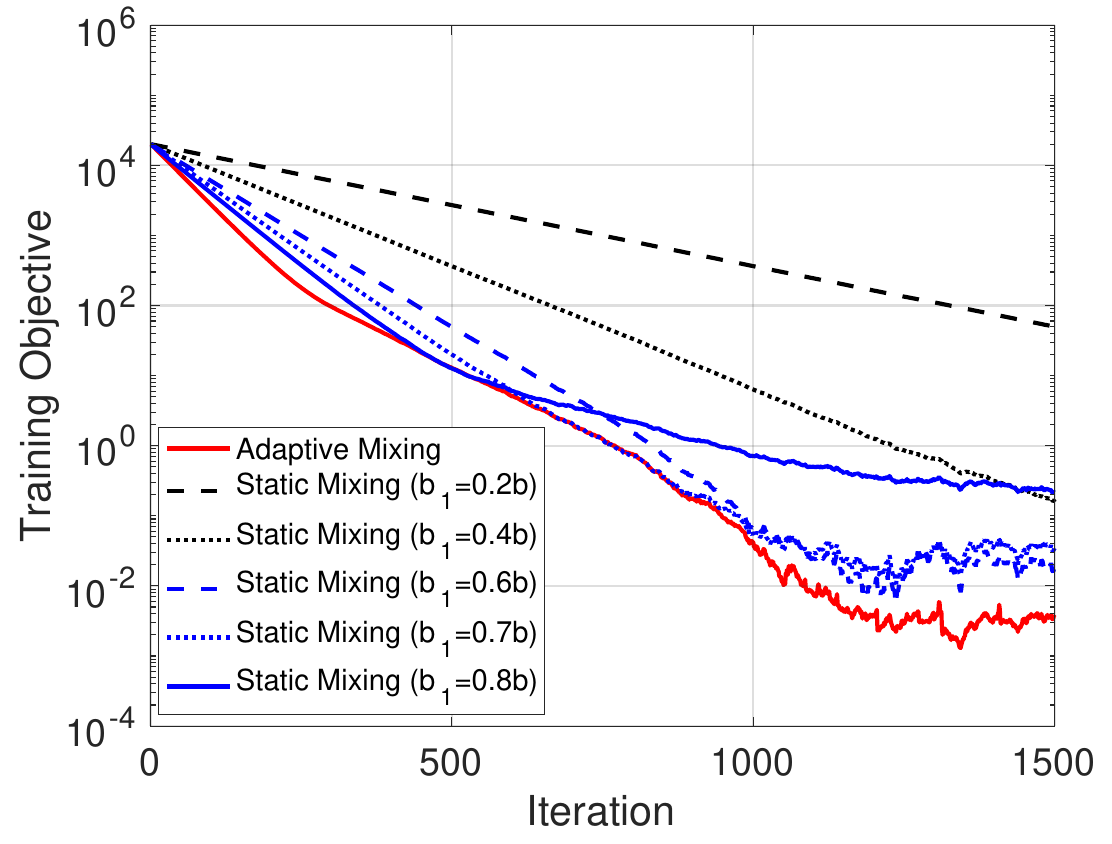}} 
\vspace{-0.2cm}
\caption{Adaptive mixing consistently outperforms static mixing in the Example~\ref{example:Adaptive}.}  
\vspace{-0.7cm}
\label{fig: ExampleAdaptiveMixingVsStatic}  
\end{center}  
\end{wrapfigure}

While such conflicts are well-documented in vision tasks, we observe they are rare in large-scale language model pretraining. In particular, as shown in Figures~\ref{fig: motivation_why_mix}, ~\ref{fig:app_mutlilingual_grad_similarity}, and \ref{fig:app_glam_grad_similarity}, task gradients are generally non conflicting in pretraining LLMs. This observation shifts the our goal: instead of mitigating conflict, we can \textbf{leverage naturally non-conflicting gradients} to improve training efficiency.

This insight renders many conflict-aware MTL methods—such as PCGrad~\citep{yu2020gradient}, AdaTask~\citep{yang2023adatask}, MGDA~\citep{sener2018multi}, and NashMTL~\citep{navon2022multi}—less effective for pre-training LLMs:
PCGrad acts only when conflicts occur (rare in LLMs) and thus performs similarly to simple \textit{Mix}. AdaTask treats different tasks as completely independent and thus does not utilizes potential cooperative nature of tasks, leading to slower convergence. More fundamentally, these methods incur substantial memory and compute overhead, making them unsuitable for large-scale models (see Appendix~\ref{app:scalability_issue_mtl} for discussions).
A notable exception is the recent work FAMO~\citep{liu2024famo}, which is designed for scalability. 

In summary, the lack of gradient conflict in LLM pretraining, combined with the limitations of existing MTL approaches, motivates the need for scalable methods that exploit non-conflicting gradient structures—rather than fixating on rare conflicts.

\subsection*{Bulding Block \#3: Adaptively Changing Mix Sampling Weights Is Necessary} \label{sec: building_blocks_3}

Prior work using the Mix sampling strategy typically relies on fixed (static) sampling weights, keeping $(b_1, \ldots, b_K)$ constant throughout training. However, dynamically adjusting batch composition can significantly enhance efficiency. We illustrate this fact with a simple example:

\begin{example}\label{example:Adaptive} Consider training on $K=2$ tasks with losses $\ell_1(\ttheta; x_1) = \frac{1}{2}(\ttheta^\top e_1)^2 + x_1^\top \ttheta$ and $\ell_2(\ttheta; x_2) = \frac{1}{2}(\ttheta^\top e_2)^2 + x_2^\top \ttheta$, where $e_1 = [1\; 0]^\top$, $e_2 = [0\; 1]^\top$, and $\ttheta \in \mathbb{R}^2$. Data for task 1 follows $x_1 \sim \cN(0, \sigma_1^2 I)$, while task 2 follows $x_2 \sim \cN(0, \sigma_2^2 I)$. The  loss for task $k$ simplifies to $\cL_k(\ttheta) = \frac{1}{2}(\ttheta^\top e_k)^2$.
Using $b_k$ samples from task~$k,$  the gradient at iteration $t$ is given by
\[
\bg_t = \frac{1}{b_1+b_2}\left(b_1e_1e_1^\top + b_2 e_2 e_2^\top\right)\ttheta_t + \bz,
\]
where $\bz \sim \cN(0, \frac{b_1 \sigma_1^2 + b_2 \sigma_2^2}{b^2} I)$ with $b = b_1 + b_2$. Updating $\ttheta_t$ via SGD, $\ttheta_{t+1} = \ttheta_t - \eta \bg_t$, we have
\begin{equation}\label{eq:ExampleExpectedLoss}
\begin{split}
    \E[\cL(\ttheta_{t+1})] &= \frac{1}{2}(1-\eta \frac{b_1}{b})^2\theta_{1,t}^2 +\frac{1}{2}(1-\eta \frac{b_2}{b})^2\theta_{2,t}^2  +\eta^2\frac{b_1\sigma_1^2+b_2\sigma_2^2}{b^2},
\end{split}
\end{equation}
where $\theta_{1,t}$ and $\theta_{2,t}$ denote the first and second component of the vector $\ttheta_t$. The derivation details of~\eqref{eq:ExampleExpectedLoss} can be found in Appendix~\ref{sec:derivationExampleExpectedLoss}. 
Letting~$w_1:=\frac{b_1}{b}$, $w_2:=\frac{b_2}{b}$,  and relaxing them  to take real values, we can optimize the mixing weights $w_1$ and $w_2$ as
\begin{equation}\label{eq:ExampleOptimalMixing}
    w_1^* = \Pi\left(\frac{ b^{-1}(\sigma_2^2-\sigma_1^2)+ \eta^{-1}(\theta_{1,t}^2-\theta_{2,t}^2)+ \theta_{2,t}^2}{\theta_{1,t}^2+\theta_{2,t}^2}\right)
\end{equation}
and $w_2^* = 1-w_1^*$ where $\Pi(\xi) = \min\{\max\{\xi,0\},1\}$ is the projection  onto~$[0,1]$. This result shows that \textbf{optimal batch composition should change over iterates} to maximize training efficiency. 

Figure~\ref{fig: ExampleAdaptiveMixingVsStatic} illustrates the superiority of the  adaptive mixing approach based on \eqref{eq:ExampleOptimalMixing} over various static mixing strategies. Moreover, the adaptive mixing strategy in this example does not require any hyperparameter tuning, while finding the best static mixing requires tuning.
\end{example}

Despite its simplicity, this example mirrors key aspects of MTL in large models:   1) The optimal solution \(\ttheta^* = 0\) minimizes all task losses simultaneously, reflecting the high expressive power of large models.  
2) Task gradients are  non-conflicting, resembling real-world gradient interactions observed in Building~Block~\#2.  Moreover, \eqref{eq:ExampleOptimalMixing} further reveals that optimal data mixing depends on \lightbluebox{(1) the gradient norm squared per task \(\|\nabla \cL_1(\ttheta)\|^2 = \theta_1^2\), \(\|\nabla \cL_2(\ttheta)\|^2 = \theta_2^2\)} and \lightorangebox{(2) gradient variance (\(\sigma_1^2\), \(\sigma_2^2\))}. 
As we will see in the next section, these factors play a crucial role in defining optimal mixing strategies for more general settings.

\section{Method}\label{sec:method}  

\subsection{PiKE: Conceptual Version}
To develop our method, we first start by quantifying gradient conflicts:

\begin{definition}\label{def:Interaction_LB}
    For a given point $\ttheta$, we say gradients are $\ulc$-conflicted (with $\ulc\geq 0$) if, for all task pairs $j,k, j\neq k$,
    \begin{align}
          -\ulc \big(\|\nabla \cL_j(\ttheta)\|^2 + \|\nabla \cL_k(\ttheta)\|^2\big) &\leq  \lin{\nabla \cL_j(\ttheta) ,\nabla \cL_k(\ttheta)}.  \nonumber 
    \end{align}
\end{definition}
The above definition is implied by a lower bound on the gradients cosine similarity. In particular, if $\frac{\lin{\nabla \cL_j(\ttheta) ,\nabla \cL_k(\ttheta)}}{\|\cL_j(\ttheta) \|\|\cL_k(\ttheta) \|} \geq -\tilde{c}$, then the gradients are $\ulc$-conflicted for $\ulc = \tilde{c}/2$. 
Therefore, experiments in Section~\ref{sec: main_building_blocks}  and Figures~\ref{fig:app_mutlilingual_grad_similarity} and~\ref{fig:app_glam_grad_similarity} in Appendix~\ref{app: cos_sim_grad_conflict} show that $\ulc$ is typically small for LLM training.

While task gradient conflict are rare in LLM training, we observed in Section~\ref{sec: main_building_blocks} that task gradients are not fully aligned either. To quantify the level of alignment, we define the following concept:

\begin{definition}\label{def:Interaction_UB}
   For a given point $\ttheta$, we say that the gradients are $\barc$-aligned (with $\barc\geq 0$) if, for all task pairs $j,k, j\neq k$,
    \begin{align}
          \lin{\nabla \cL_j(\ttheta) ,\nabla \cL_k(\ttheta)}  \leq \barc \|\nabla \cL_j(\ttheta)\|_2 \|\nabla \cL_k(\ttheta)\|_2. \nonumber
    \end{align}
\end{definition}
Notice that~\Cref{def:Interaction_LB} and \ref{def:Interaction_UB} always hold for \(\barc = 1\) and \(\ulc = 1/2\). However, smaller values allow for more refined analysis and more desirable properties. Notably, when both \(\barc\) and \(\ulc\) are small (as we observed in our experiments), the value of \(\|\nabla \cL(\ttheta)\|\) is small if and only if  \(\|\nabla \cL_k(\ttheta)\|\) is small for all \( k \) (see Lemma~\ref{le: correlation_of_losses} in Appendix~\ref{app: theory}). To proceed further, we make the following standard assumption:

\begin{assumption}\label{as:assumption1}
Per-task gradients are $L$-Lipschitz, unbiased, and have bounded variance, i.e., $\forall k:$
\begin{align}
    & \|\nabla \cL_k(\ttheta_1) - \nabla \cL_k(\ttheta_2)\|\leq L\|\ttheta_1 - \ttheta_2\|, \;\;\forall \ttheta_1,\ttheta_2\\
    & \E_{x\sim \cD_k}[\nabla \ell_k(\ttheta; x)] = \nabla \cL_k(\ttheta), \;\;\forall \ttheta\\
    & \E_{x\sim \cD_k}[\|\nabla \ell_k(\ttheta;x) - \nabla\cL_k(\ttheta)\|^2] \leq \sigma_k^2,\;\;\forall \ttheta
\end{align}
\end{assumption}

Using a mix batch with \( b_k \) samples per task~$k$, the estimated gradient follows \eqref{eq: mix_framework}. The next theorem characterizes the descent amount in one step of SGD  under low conflict conditions:

\begin{theorem}
\label{thm:DescentMainBody}
Assume Assumption~\ref{as:assumption1} holds and the gradients at $\ttheta_t$ are $\ulc$-conflicted and $\barc$-aligned, with $\ulc < \frac{1}{K - 2 + b/b_k}$ for all $k$. If gradients are computed using the \textit{Mix} strategy~\eqref{eq: mix_framework}, then:
    \begin{align} \label{eq: theorem1}
        \E[\cL(\ttheta_{t+1})] &\leq \cL(\ttheta_t) + \sum^K_{k=1} b_k\Big(-\frac{\eta}{b}\beta\|\nabla\cL_k(\ttheta_t)\|^2 + \frac{L\eta^2}{2b^2}\sigma_k^2\Big)  + \sum^K_{k=1} b_k^2 \frac{L\eta^2}{2b^2}\gamma \|\nabla \cL_k(\ttheta_t)\|^2 
    \end{align}
   for SGD update $\ttheta_{t+1} = \ttheta_t - \eta \bg_t$. Here, $\beta \triangleq \min_{k} (1+\ulc(-K+2-\frac{b}{b_k}))$, $\gamma \triangleq 1 + \barc (K-1)$, and the expectation is over batch sampling randomness under the mix strategy $(b_1, \ldots, b_K)$.
\end{theorem}
A formal proof is provided in Theorem~\ref{thm: formal_descent_lemma} (Appendix~\ref{app: theory}). Theorem~\ref{thm:DescentMainBody} provides an upper-bound on the decrease in the objective depending on the mix batch composition~$(b_1,\ldots,b_K)$. As we show in Theorem~\ref{thm:DescentLowerBound} in the appendix, this upperbound is tight in the non-conflicting, non-aligned  regime of $\barc=\ulc=0$ (which is a good approximation for pre-training LLMs according to our experiments). 

According to Theorem~\ref{thm:DescentMainBody}, to maximize descent in Mix sampling, we need to minimize the RHS of~\eqref{eq: theorem1}. Thus,  relaxing  \( w_k = b_k / b \) to continuous values, we need to solve:  
\begin{equation} \label{eq:relax_problem}
    \min_{w_1,\ldots,w_K\geq0} \; \sum^K_{k=1}  w_k \lambda_k + \frac{1}{2} w_k^2\kappa_k 
    \;\; \textrm{s.t.} \;\; 
    \sum_{k=1}^Kw_k = 1
\end{equation}
where $\lambda_k\triangleq -\eta\beta\|\nabla\cL_k(\ttheta)\|^2 + \frac{L\eta^2}{2b} \sigma_k^2$ and $\kappa_k\triangleq L\eta^2\gamma\|\nabla\cL_k(\ttheta)\|^2$. Using KKT conditions, the optimal solution is given by 
$
w_k^* = \max\left\{0, -\frac{\mu + \lambda_k}{\kappa_k}\right\},
$
where $\mu$ is chosen such that $\sum_{k=1}^K w_k^* = 1$ (see Lemma~\ref{le:optimal_w_for_relax_problem},  Appendix~\ref{app: theory}). 
This   leads to the conceptual version of PiKE (Positive gradient Interactions-based K-task weight Estimator),  summarized in Algorithm~\ref{alg: Basic PiKE} in Appendix~\ref{sec:ConceptualPiKE}.

The conceptual version of PiKE (Algorithm~\ref{alg: Basic PiKE}) adaptively adjusts sampling weights. This adaptive adjustment makes the stochastic gradients biased, i.e., $\E [\bg_t] \neq \nabla \cL(\ttheta_t)$. Due to this introduced bias, the classical convergence results of SGD can no longer be applied. The following theorem establishes the convergence of  conceptual PiKE:

\begin{theorem}%
\label{thm:IterationComplexityConceptualPiKe}
Suppose the assumptions in Theorem~\ref{thm:DescentMainBody} hold and the Conceptual PiKE Algorithm (Algorithm~\ref{alg: Basic PiKE} in the Appendix) initialized at $\ttheta_0$ with the SGD optimizer in Step 10 of the algorithm. Let $\Delta_L = \cL(\ttheta_0) - \min_{\ttheta}\cL(\theta)$ and $\sigma_{\max} = \max_k \sigma_k$. Suppose $\delta>0$ is a given constant and the stepsize $\eta \leq \frac{\beta \delta}{L\sigma_{\max}^2/b + L\gamma \delta}$. Then, after $T = \frac{2 \Delta_L}{\eta \beta \delta}$ iterations, Algorithm~\ref{alg: Basic PiKE} finds a point $\bar{\ttheta}$ such that 
\begin{equation}
    \label{eq:boundedNormGrad}
    \mathbb{E} \|\nabla \cL_k(\bar{\theta})\|^2 \leq \delta,\quad \forall k=1,\ldots, K.
\end{equation}
Particularly, if we choose $\eta = \frac{\beta \delta}{L\sigma_{\max}^2/b + L\eta \delta}$, then the Conceptual PiKE algorithm requires at most 
$
\bar{T} = \frac{2L\Delta_L(\sigma_{\max}^2/b + \gamma \delta)}{\delta^2 \beta^2}
$
iterations to find a point satisfying~\eqref{eq:boundedNormGrad}.
\end{theorem}
This theorem (restated as Theorem~\ref{thm:app_IterationComplexityConceptualPiKe}) is proved in Appendix~\ref{app: theory}. 

\begin{remark} 
Theorem~\ref{thm:IterationComplexityConceptualPiKe} guarantees that, given enough iterations, all tasks are learned jointly, i.e., the gradients of all  losses become small. Moreover, the convergence rate is $T = O(1/\delta^2)$, which matches the optimal iteration complexity for smooth, nonconvex stochastic optimization.
\end{remark}

\begin{remark}
The Conceptual PiKE algorithm maximizes the expected decrease at each iteration by finding the optimal mix batching strategy. This property leads to improved iteration complexity bound than (static) uniform mix batching, as discussed in \Cref{app: PiKEvsUniform}, when $\ulc$ and $\barc$ are small.
\end{remark}

\normalsize

\subsection{PiKE: Simplified Computationally Efficient Version}  
\label{sec:PiKe}  

Solving \eqref{eq:relax_problem} requires estimating \lightbluebox{$\{ \| \nabla \mathcal{L}_k (\boldsymbol{\theta}_t) \|^2 \}_{k=1}^K$} and \lightorangebox{$\{ \sigma_k \}_{k=1}^K$}; however, per-iteration estimation of these terms with sufficient accuracy often necessitates large batch computations, thereby impeding convergence. To speed up the algorithm, we can update these estimates every \(T_0\) iterations. However, this can cause abrupt changes in \((w_1,\ldots, w_K)\), leading to instability, especially with optimizers like Adam, where sudden shifts may disrupt momentum estimates.  
To mitigate this, we update \((w_1, \dots, w_K)\) using a single mirror descent step on \eqref{eq:relax_problem}, ensuring gradual adjustments:  

\vspace{-0.3cm}

{\small  
\begin{align}  
    w_k \gets w_k \exp\left(  
    \alpha \eta (\beta-L\eta \gamma w_k) \stackrel{\text{\tiny gradient norm}}{\lightbluebox{ $\|\nabla \cL_k(\ttheta)\|^2$}}
    - \frac{\alpha L \eta^2}{2b} \stackrel{\text{\tiny gradient variance}}{\lightorangebox{ $\quad \sigma_k^2 \quad $ }}  
    \right) \nonumber  
\end{align}  
}  

\vspace{-0.3cm}

\noindent followed by normalization: \( \bw \gets \bw / \|\bw\|_1 \), where \(\alpha\) is the mirror descent step size.  

Even after the above simplifications, fine-tuning \( L, \gamma, \alpha, \) and \( \beta \) can be challenging in practice. We simplify this finetuning by noting two observations:  
1) The coefficient in front of \( \sigma_k^2 \) is constant, independent of \( w_k \).  
2) For small \( \eta \) and \( w_k < 1 \), the coefficient in front of \( \|\nabla \cL_k (\ttheta)\| \) remains nearly constant:  
$  
  \alpha \eta (\beta-L\eta \gamma w_k) \approx \alpha \eta \beta.
$  
Therefore, in our practical implementation, PiKE employs tunable constant coefficients for terms related to task-specific gradient variance and gradient norms, which simplifies its application. 
The resulting algorithm is detailed in Algorithm~\ref{alg: main}. 

\begin{remark}
   The inclusion of the variance-related term, $\sigma_k^2$,  is crucial in PiKE; its omission would cause PiKE to disproportionately prioritize tasks with currently large losses even if their gradients are very noisy (high variance). 
Our ablation study, presented in Appendix~\ref{app:pike-variance-magnitude-estimation-details}, underscores the importance of this term for achieving robust and balanced multi-task performance. 
\end{remark}

\begin{remark}
PiKE updates sampling weights only every $T_0$ steps. In our experiments, this additional step incurs negligible memory and computational overhead (see Appendix~\ref{app:pike-training-overhead}).
\end{remark}

\begin{algorithm}[!tb] 
    \begin{algorithmic}[1]
        \STATE {{\bfseries Input:} $\ttheta$, $T_0$, total batch size $b$, task $k$ dataset $\cD_k$, hyperparameters $\zeta_1$ and $\zeta_2$, prior weights~$\bw'$ } 
	\STATE {{\bfseries Initialize:} $w_k \gets 1/K$ or $w_k \gets w_k'$ } 
		\FOR{$t=0,1, \dots$}
                \IF{$t \mod T_0 = 0$}
                    \STATE Estimate $\|\nabla \cL_k(\ttheta_t)\|^2$ and $\sigma_k^2$ for every $k$
           
                    \STATE $w_k \gets w_k \exp\bigg(\zeta_1  \lightbluebox{$\|\nabla \cL_k(\ttheta_t)\|^2$} - \frac{\zeta_2}{2b} \lightorangebox{$\sigma_k^2$} \bigg)$
                    \STATE  $\bw \gets \bw /\|\bw\|_1$
                    \STATE $(b_1,\ldots, b_K) \gets \textrm{round}(b (w_1,\ldots, w_K))$
                \ENDIF
		        \STATE Sample $b_k$ data points from each task~$k$
                \STATE Compute the gradient $\bg$ using the estimates samples
                \STATE{Update: $\ttheta_{t+1} \gets \textrm{Optimizer} (\eta, \ttheta_t,\bg)$} 
		\ENDFOR
    \end{algorithmic}
    \caption{PiKE: Positive gradient Interaction-based K-task weights Estimator}
    \label{alg: main}
\end{algorithm}

\vspace{0.2cm}
\subsection{Balanced-PiKE: Balanced Learning Across Different Tasks}  
\label{sec:fair-pike}

While Algorithm~\ref{alg: main} optimizes the average loss across tasks~\eqref{eq:prob_def}, this objective can lead to unbalanced learning. In practice, some tasks may dominate due to larger datasets or easier loss reduction, causing others to be under-optimized. This imbalance is problematic when certain domains (e.g., code, math) are more important to user, or when balanced performance across all tasks is desired. To address this shortcoming, we consider a balance-promoting objective based on \textit{tilted empirical risk minimization}~\citep{li2020tilted}, also known as the $\alpha$-\textit{fairness utility}~\citep{mo2000fair}:
\begin{equation}\label{eq:fairness_def}  
   \min_{\ttheta} \;\; \widetilde{\cL}(\tau ; \ttheta) := \frac{1}{\tau} \log \left(\sum^K_{k=1} e^{\tau\cL_k(\ttheta)}\right).
\end{equation}

This objective interpolates between average loss (\(\tau \to 0\)) and worst-case loss (\(\tau \to \infty\)), allowing users to control the trade-off between efficiency and fairness. Moderate values of $\tau$ ($\tau \approx 1\sim 3$ in our experiments) encourage balanced learning without sacrificing overall performance.

To leverage our PiKE developments in solving \eqref{eq:fairness_def}, we need to connect it to our intital objective in~\eqref{eq:prob_def}. The following lemma achieves this through the use of Fenchel duality \citep{rockafellar2015convex}:

\begin{lemma}\label{lemma:MinToMinMax}
    Assume $\cL_k(\ttheta)>0, \;\forall k$ and let $0<\tau \in \mathbb{R}$. Then \eqref{eq:fairness_def} is equivalent to solving
    \begin{align}
    \min_{\ttheta} \quad \max_{\substack{\mathbf{y} \in \mathbb{R}_{+}^K \\ \sum_{k=1}^K y_k=\tau}} \sum^K_{k=1} y_k\cL_k(\ttheta) - \sum^K_{k=1} \frac{y_k}{\tau} \log \left(\frac{y_k}{\tau} \right). \label{eq:MinMaxFairPiKE}
    \end{align}  
    Moreover, for any fixed $\ttheta$, the inner maximization problem is maximized at
    $y_k^\star = \frac{\tau e^{\tau\cL_k(\ttheta)}}{\sum^K_{j=1} e^{\tau\cL_j(\ttheta)}}, \; \forall k.$
\end{lemma}

This lemma, which is proved in~\Cref{app:fairness_theory}, provides a natural alternative minimization algorithm for solving~\eqref{eq:fairness_def}: 
Fixing \(\mathbf{y}\), \eqref{eq:MinMaxFairPiKE} reduces to a weighted minimization over tasks, where \textit{regular PiKE sampling} with proper weights $y_k$ in front of each loss can be applied to determine the optimal mixing strategy. On the other hand, fixing $\ttheta$, the optimal solution $y_k^\star$ can be computed according to \Cref{lemma:MinToMinMax}.
This leads to \textit{Balanced-PiKE} algorithm, described in Appendix~\ref{app:pike-fairness}, which balances overall loss minimization and balanced/fair learning of all tasks. Although prior MTL literature explored fair learning~\citep{ban2024fair, navon2022multi}, such methods often grapple with scalability limitations, a critical aspect where PiKE is designed to excel through efficient large-scale performance.

\begin{table*}[!tb]
\large
\centering
\caption{We report the perplexities (lower the better) on the validation split of multilingual C4 datasets. We also report the accuracies (\%, higher the better) of different models on HellaSwag and its corresponding translated version.~\textbf{Bolding} indicates the best model in the task; $\ols{\text{Metrics}}$ means the average across different tasks. Additional results can be found in Table~\ref{tab:main-table-multilingual}.}
\begin{adjustbox}{max width=\textwidth}
\begin{tabular}{lcccc|cccc}
\toprule
            &               & C4 (en)       & C4 (hi)       & C4 (de) &               & HellaSwag (en) & HellaSwag (hi) & HellaSwag (de) \\ \cmidrule(lr){3-5} \cmidrule(l){7-9} 
 &
  $\ols{\text{Perplexity} \downarrow}$ &
  Perplexity $\downarrow$ &
  Perplexity $\downarrow$ &
  Perplexity $\downarrow$ &
  $\ols{\text{Accuracy}(\%) \uparrow}$ &
  0-shot $\uparrow$ &
  0-shot  $\uparrow$ &
  0-shot $\uparrow$ \\ \midrule
\multicolumn{9}{l}{\textbf{C4 (en), C4 (hi), and C4 (de) datasets, GPT-2 large style, 1B params, 36 Layers default, 120K training steps}}                     \\
Mix & 8.29 & 11.13 & \textbf{4.45} & 9.29 & 27.5 & 28.1 & 27.1 & 27.6 \\
Round-Robin & 8.41 & 11.31 & 4.97 & 9.46 & 26.5 & 27.6 & 26.7 & 26.3 \\
Random & 8.48 & 11.38 & 4.54 & 9.55 & 26.6 & 27.0 & 26.9 & 26.1 \\ \midrule
FAMO~\cite{liu2024famo}  & 8.25 & 11.04 & 4.48 & 9.23 & 27.2 & 27.3 & 26.9 & 27.3 \\
ADO~\cite{jiang2024adaptive} & 8.30 & 11.12 & 4.45 & 9.31 & 27.5 & 27.7 & \textbf{27.5} & 27.2  \\
PiKE & 9.56 & \textbf{9.49} & 5.32 & 13.87 & 28.7 & \textbf{33.0} & 27.2 & 26.2 \\ 
Balanced-PiKE ($\tau=1$) & 8.29 & 11.12 & 4.46 & 9.31 & 27.9 & 28.3 & 27.4 & 28.0 \\
Balanced-PiKE ($\tau=3$) & \textbf{8.18} & 10.14 & 4.93 & 9.49 & \textbf{28.9} & 31.3 & 27.3 & 28.1 \\
Balanced-PiKE ($\tau=5$) & 8.42 & 10.02 & 6.30 & \textbf{8.94} & \textbf{28.9} & 31.2 & 26.9 & \textbf{28.6} \\
\bottomrule
\end{tabular}
\end{adjustbox}
\label{tab:partial_main-table-multilingual}
\end{table*}

\begin{table*}[!tb]
\centering

\caption{We report perplexity (lower is better) on the validation split of the GLaM datasets, averaging perplexities across six domains. We also compare the accuracies (\%, higher the better) of different models on four different Q/A tasks. PiKE (Uniform) means PiKE using initial uniform sampling weights and PiKE (GLaM) means PiKE using GLaM tuned weights as initial weights. \textbf{Bolding} indicates the best model in the task, \ul{underlining} indicates PiKE beating Mix, Round-Robin, Random methods. Additional detailed (per-domain) results with different model sizes can be found in Table~\ref{tab:main-table-glam}.}

{\small
\begin{adjustbox}{max width=0.8\textwidth}
\begin{tabular}{lc|ccccc}
\toprule
                 & GLaM              &                      & ArcE                 & CSQA            & HellaSwag                 & PIQA                 \\ \cmidrule(l){4-7} 
 &
  $\ols{\text{Perplexity} \downarrow } $ &
  $\ols{\text{Accuracy} (\%) \uparrow}$ &
  7-shot $\uparrow$ &
  7-shot $\uparrow$ &
  7-shot  $\uparrow$ &
  7-shot $\uparrow$ \\ \midrule
\multicolumn{7}{l}{\textbf{Six domains of GLaM dataset, GPT-2 large style, 750M params, 36 layers default}}                       \\
Mix &\textbf{12.77} & 46.4 & 47.2 & 39.6 & 37.9 & 60.9        \\
Round-Robin &12.98 & 44.3 & 43.5 & 36.7 & 36.8 & 60.3 \\
Random &12.99 & 42.7 & 41.7 & 34.2 & 36.6 & 58.2                 \\ \midrule
FAMO~\cite{liu2024famo} & 13.25 & 45.0 & 43.7 & 40.0  & 36.4 & 59.8         \\
ADO~\cite{jiang2024adaptive} & \textbf{12.77} & 45.9 & 45.5 & 38.7   & 38.1  & 61.1           \\
GLaM~\cite{du2022glam} &13.20 & 45.3 &46.9 & 39.8 & 38.0 & 56.4            \\
DoReMi~\cite{xie2024doremi} &13.25 & 46.5 & 48.6 & 40.1 & 37.5 & 59.6         \\
PiKE (Uniform) &13.22 & \ul{47.6} &\ul{49.6} & \ul{43.2} & 37.2 &60.4   \\
PiKE (GLaM) &13.35 & \ul{48.1} & \ul{\textbf{49.8}} & \ul{\textbf{43.5}} & \ul{38.0} & \ul{61.2} \\
Balanced-PiKE ($\tau=1$) & 13.21 & \ul{47.5} & \ul{48.8} & \ul{42.5} & 37.6 & \ul{61.2} \\
Balanced-PiKE ($\tau=3$) & 13.26  & \ul{47.2} & \ul{48.8}  & \ul{41.5}   & 37.2 & \ul{61.3}  \\
Balanced-PiKE ($\tau=5$) & 13.19 & \ul{\textbf{48.2}} & \ul{49.3} & \ul{42.6} & \ul{\textbf{38.5}} & \ul{\textbf{62.4}} \\
\bottomrule
\vspace{0.1cm}
\end{tabular}
\end{adjustbox}
}
\label{tab:partial_main-table-glam}
\end{table*}

\section{Experiments}\label{sec:exp}  
We evaluate PiKE in two multitask pretraining scenarios:  
1) \textit{Pretraining language models on multilingual mC4 dataset}~\citep{xue2020mt5}, a dataset covering diverse languages from Common Crawl corpus.  
2) \textit{Pretraining language models on the   GLaM dataset}~\citep{du2022glam}, an English dataset spanning six domains.  
As we will show, across multiple model sizes (110M, 270M, 750M, and 1B parameters), PiKE consistently outperforms all existing methods, including heuristic or static data mixing, multitask optimization, and adaptive data mixture approaches. We first start by explaining our setup.

\subsection{Experiment Setup}  

\vspace{-0.1cm}

\textbf{Baselines:}  
We evaluate a range of sampling strategies: (1) \textit{(Uniform) Mix}, (2) \textit{Round-Robin}, (3) \textit{Random}, (4) \textit{FAMO}~\citep{liu2024famo}, (5) \textit{ADO}~\citep{jiang2024adaptive}, (6) \textit{GLaM}~\citep{du2022glam}, (7) \textit{DoReMi}~\citep{xie2024doremi},  (8) \textit{PiKE}, and (9) \textit{Balanced-PiKE}. Among these, DoReMi, GLaM, and ADO are specifically designed for LLM pretraining. We also include FAMO, a recent multitask learning (MTL) method, because it is scalable to large model sizes and has demonstrated competitive performance—serving as a useful comparison point for our work. DoReMi estimates task weights by training a small proxy model, while GLaM assigns static weights based on downstream performance from smaller models. However, since both methods report weights only for the GLaM dataset and do not provide configurations for multilingual C4, we exclude them from our multilingual experiments. In contrast to these static methods, \textit{PiKE} dynamically updates task sampling weights during training using gradient information, enabling adaptive optimization throughout training. PiKE is scalable to both large models and many tasks, with minimal overhead, as detailed in Appendix~\ref{app:pike-training-overhead}.

\textbf{Datasets:}  
For multilingual experiments, we use mC4~\citep{xue2020mt5}, focusing on English (en), Hindi (hi), and German (de). An overview of these datasets is provided in Table~\ref{tab: mc4-dataset-overview}. For GLaM-based experiments, we use the six-domain GLaM dataset~\citep{du2022glam}. Additional details are presented in Table~\ref{tab: glam-dataset-overview}.  

\textbf{Evaluation:}  
Perplexity is measured on held-out validation data. Downstream evaluation follows the OLMES suite~\citep{gu2024olmes}. For multilingual downstream tasks, we use multilingual HellaSwag~\citep{dac2023okapi}, covering 26 languages. For  models trained on GLaM, we evaluate on downstream tasks ARC-Easy~\citep{clark2018think}, CommonsenseQA~\citep{talmor2018commonsenseqa}, PIQA~\citep{bisk2019reasoning}, and HellaSwag~\citep{zellers2019hellaswag}. HellaSwag and ArcE tasks have 4 choices, CSQA has 5 choices, and PIQA
has 2 choices.

Further details on our experimental setup and evaluation are in Appendix~\ref{app:experiment_setup}.

\subsection{Main Observations on Multilingual Pretraining Experiments}   
Table~\ref{tab:partial_main-table-multilingual} presents results for pretraining a 1B multilingual GPT-2 model~\citep{radford2019language} on English, Hindi, and German. Mix batching outperforms Round-Robin and Random strategies, supporting our analysis in Section~\ref{sec: building_blocks_1} and justifying our focus on Mix as the foundation for PiKE. Additional results with different model sizes (270M and 1B) and language settings are reported in Table~\ref{tab:main-table-multilingual}. Our main observation is that
\textit{PiKE and its Balanced variant  achieve the highest average downstream accuracy} across all language settings and model sizes, demonstrating their effectiveness for multilingual pretraining. 

We also observe that \textit{Balanced-PiKE promotes more fair learning across tasks}. In particular, we pre-trained 1B models using Balanced-PiKE with different values of parameter \(\tau \in \{1, 3, 5\}\). As \(\tau\) increases, task losses become more uniform, reflecting improved balanced learning. At \(\tau = 5\), perplexity becomes more balanced across languages, while \(\tau = 3\) offers the best trade-off—achieving both the lowest perplexity and highest downstream accuracy. These results highlight the importance of incorporating fairness/balanced learning into data mixing strategies during pretraining.

\subsection{Main Observations on Pretraining Experiments with GLaM Datasets}  
Table~\ref{tab:partial_main-table-glam} shows results for pretraining a 750M GPT-2 model on the GLaM dataset. Additional results with both 110M and 750M models across six domains are provided in Table~\ref{tab:main-table-glam}.
Across both 110M and 750M model sizes, \textit{PiKE consistently outperforms} DoReMi, GLaM, and Mix in downstream accuracy. For the 750M model, PiKE improves average accuracy by \textbf{3.4\%} over DoReMi, \textbf{6.2\%} over GLaM, \textbf{7.1\%} over FAMO, and \textbf{4.8\%} over ADO. In the 110M setting, PiKE achieves \textbf{37.8\%} accuracy, exceeding DoReMi (\textbf{36.0\%}), GLaM (\textbf{35.3\%}), and FAMO (\textbf{35.9\%}). Unlike DoReMi, which requires a separate proxy model, or GLaM, which depends on tuning with smaller models, PiKE delivers these gains with negligible additional overhead.

\textit{PiKE benefits from apriori downstream-tuned weights.}  
We evaluate PiKE with two initializations: (1) uniform weights $b_k = b/K$ and (2) GLaM-tuned weights. In both small and large GPT-2 configurations, PiKE benefits from utilizing already fine tuned  weights as initialization, achieving \textbf{48.1\%} accuracy with GLaM-tuned weights vs. \textbf{47.6\%} with uniform initialization. This shows that PiKE can effectively leverage pre-existing fine-tuned weights while still outperforming other methods with uniform initialization.  

\textit{Mixing datasets improves language model generalization.}  
We compare models trained on individual domains to those trained on mixed-domain datasets. Table~\ref{tab:main-table-glam} shows that single-domain training underperforms compared to mixed-domain training, even with simple Mix sampling. This reinforces the importance of diverse data for pretraining and aligns with prior work~\citep{liu2024regmix, hoffmann2022empirical}.  

\textit{Perplexity versus downstream performance.}  
Table~\ref{tab:main-table-glam} reveals that validation perplexity does not always align with downstream performance. For instance, while Mix sampling yields lower perplexity in 750M models, PiKE achieves better downstream accuracy. This aligns with prior findings~\citep{gao2025metadata, tay2021scale, liu2023same, wettig2024qurating}, suggesting that perplexity alone is not a reliable performance metric.

\section{Conclusion}
In this work, we introduced \textit{PiKE}, an adaptive data mixing algorithm for multitask learning that dynamically adjusts task sampling based on observed gradient interactions. Unlike prior methods that aim to resolve gradient conflicts, PiKE exploits the predominantly non-conflicting gradients seen in large-scale language model pretraining. We provided theoretical analysis and showed, through extensive experiments, that PiKE improves both convergence speed and downstream performance across multilingual and multi-domain settings.
To promote balanced task learning, we extended PiKE with a fairness-aware objective, resulting in \textit{Balanced-PiKE}, which reduces task-level performance gaps without sacrificing overall accuracy.

One limitation of PiKE is its lack of sensitivity to dataset size when assigning sampling weights. Future work could incorporate data abundance or downstream performance feedback into the sampling strategy. Additionally, extending PiKE to other domains beyond language modeling remains a promising direction for further research.

\bibliography{reference}
\bibliographystyle{abbrvnat}

\clearpage
\appendix

\section{Related Work} \label{sec: related}

\textbf{\small Data Curation and Selection.} The effectiveness of language models heavily depends on the quality of the pre-training corpus. Consequently, significant efforts have been made to enhance pre-training data. These efforts include heuristic-based filtering~\citep{raffel2020exploring, rae2021scaling, laurenccon2022bigscience, penedo2023refinedweb, soldaini2024dolma} and deduplication~\citep{abbas2023semdedup, lee2021deduplicating, chowdhery2022palm, dubey2024llama}. Recently, \cite{vo2024automatic} proposed an automated method for constructing large, diverse, and balanced datasets for self-supervised learning by applying hierarchical k-means clustering. \cite{sachdeva2024train} introduced techniques that leverage instruction-tuned models to assess and select high-quality training examples, along with density sampling to ensure diverse data coverage by modeling the data distribution. Additionally, \cite{guu2023simfluence} simulated training runs to model the non-additive effects of individual training examples, enabling the analysis of their influence on a model's predictions.

\textbf{\small Multitask Learning Optimization}  
The approach most closely related to our method is multitask learning (MTL) optimization, which modifies gradient updates to mitigate gradient conflicts—situations where task gradients point in opposing directions, slowing down optimization~\citep{vandenhende2021multi, yu2020gradient}. The Multiple Gradient Descent Algorithm (MGDA)~\citep{desideri2012multiple, sener2018multi} updates the model by optimizing the worst improvement across all tasks, aiming for equal descent in task losses. Projected Gradient Descent (PCGrad)~\citep{yu2020gradient} modifies task gradients by iteratively removing conflicting components in a randomized order, ensuring that updates do not interfere destructively across tasks. Conflict-Averse Gradient Descent (CAGRAD)~\citep{liu2021conflict} optimizes for the worst task improvement while ensuring a decrease in the average loss. NASHMTL~\citep{navon2022multi} determines gradient directions by solving a bargaining game that maximizes the sum of log utility functions. While these methods improve performance, they introduce significant computational and memory overhead, making them impractical for large-scale models with numerous tasks~\citep{xin2022current}. Similar challenges exist in AdaTask~\citep{yang2023adatask}, which improves multitask learning by balancing parameter updates using task-wise adaptive learning rates, mitigating task dominance, and enhancing overall performance. Unlike previous approches that requires  requiring \(O(K)\) storage for task gradients (e.g. PCGrad) or optimizer states (e.g. AdaTask), FAMO~\citep{liu2024famo} balances task loss reductions efficiently using \(O(1)\) space and time. However, these methods fail to exploit the~\textit{non-conflicting} interactions among tasks, focusing instead on resolving conflicts that seldom arise. This highlights the need for a new approach that actively leverages lack of gradient conflicts to enhance training efficiency. 

Another line of work focuses on adjusting the domain mixture to improve data efficiency during training~\citep{xie2024doremi, xia2023sheared, jiang2024adaptive}. However, these methods require a target loss for optimization, which has been shown to not always correlate with downstream performance~\citep{tay2021scale, liu2023same, wettig2024qurating}. In contrast, our method leverages the absence of gradient conflict and the presence of positive gradient interactions between tasks or domains. This approach provides a more reliable and effective way to enhance the final model's performance.

\section{The Scalability Problem of Existing MTL Methods}\label{app:scalability_issue_mtl}

Many existing multitask learning (MTL) methods~\citep{desideri2012multiple, sener2018multi, wang2020gradient, yu2020gradient, liu2021conflict, navon2022multi, yang2023adatask, ban2024fair} require computing and storing all $K$ task-specific gradients at each training iteration, where $K$ is the number of tasks. While exceptions like FAMO~\citep{liu2024famo} are designed for efficient scaling, the aforementioned general approach typically leads to $O(Kd)$ space complexity for storing gradients and $O(Kd)$ time complexity per iteration for their computation, where $d$ denotes the number of model parameters. This cost is further exacerbated when these methods also involve solving auxiliary optimization problems to combine or re-weight gradients~\citep{xin2022current}. In stark contrast, when the overall loss is a simple average of task-specific losses (e.g., $\mathcal{L} = \frac{1}{K} \sum_{k=1}^K \mathcal{L}_k$), its gradient $\nabla \mathcal{L}$ can be computed via a single backward pass. This results in $O(d)$ space and time complexity for the gradient computation, with the task-aggregation step being effectively independent of $K$.

While early MTL research often focused on relatively small-scale vision models or language models with fewer than 100 million parameters, modern language models frequently exceed 100 billion parameters (100B). Storing the gradients or weights for such a model (assuming 32-bit floating-point precision) requires approximately 400\,GB of memory. This capacity already surpasses that of a single high-performance GPU like the NVIDIA H100 (80\,GB), necessitating at least five such GPUs for storing just one set of gradients or weights.

Applying typical prior MTL methods in this large-scale context becomes impractical. Storing $K$ full gradients for a 100B parameter model would demand approximately $400K$\,GB of memory, equivalent to $5K$ H100 GPUs. This scaling of memory and computational requirements—at least linear with the number of tasks $K$ for gradient operations and often compounded by super-linear costs from auxiliary optimization steps—renders naive applications of many existing MTL methods infeasible for current large language models. Indeed, even if storing per-task gradients were feasible, \citet{xin2022current} reported a significant reduction in training throughput (steps per second) for such MTL techniques even on small-scale vision models, let alone for modern LLMs (e.g., those with 1B+ parameters). Collectively, these challenges highlight the pressing need for novel MTL methodologies engineered for large-scale pre-training, demanding both exceptional scalability and minimal additional computational and memory overhead.

\section{Discussion of Gradient Conflicts} \label{app: gradient_conflicts}
GradVaccine~\citep{wang2020gradient} made a similar observation regarding low gradient conflicts among task gradients in multilingual BERT (178M parameters). We extend this finding to substantially larger autoregressive, decoder-only models (up to 1B parameters), which are more characteristic of current large-scale language modeling paradigms.

Our experiments demonstrate that task gradients in large-scale models indeed exhibit minimal conflicts. To illustrate, we conduct two pre-training experiments: (i) a 1B parameter GPT-2-style model~\citep{radford2019language} on the multilingual mC4 dataset~\citep{xue2020mt5} (covering six languages: English, Hindi, German, Chinese, French, and Arabic), and (ii) a 750M parameter model on The GLaM dataset~\citep{du2022glam} (English text from six diverse domains). Experimental details are provided in Appendix~\ref{app:experiment_setup}. Figures~\ref{fig: motivation_why_mix} and~\ref{fig:app_mutlilingual_grad_similarity} depict cosine similarity trends for task gradients, revealing several key observations: 1) Gradient similarity is initially high but generally decreases as training progresses. 2) In multilingual settings, gradient similarity correlates with linguistic proximity (e.g., English-German gradients align more closely), whereas gradients from GLaM's diverse domains exhibit more uniform positive alignment. 3) Task gradients rarely conflict—multilingual cosine similarities seldom drop below -0.1, and GLaM domain gradients remain predominantly positive.

\section{PiKE: Conceptual Version}
\label{sec:ConceptualPiKE}
Here, we present the conceptual (basic) version of PiKE. As discussed in the main text, this approach lacks computational efficiency due to the frequent estimation of the norm and the variance of the per-task gradient.

\begin{algorithm}[H] 
    \begin{algorithmic}[1]
        \STATE {{\bfseries Input:} $\ttheta$, total batch size $b$, stepsize $\eta$, task $k$ dataset $\cD_k$, constants $\beta$, $L$, $\gamma$, and prior weights~$\bw'$ } 
	\STATE {{\bfseries Initialize:} $w_k \gets 1/K$ or $w_k \gets w_k', \forall k$ } 
		\FOR{$t=0,1, \dots$}
                \STATE Estimate $\|\nabla \cL_k(\ttheta_t)\|^2$ and $\sigma_k^2$ for every $k$
                \STATE Compute $\lambda_k\triangleq -\eta\beta\|\nabla\cL_k(\ttheta_t)\|^2 + \frac{L\eta^2}{2b} \sigma_k^2$ and $\kappa_k\triangleq L\eta^2\gamma\|\nabla\cL_k(\ttheta_t)\|^2$
                \STATE set $w_k^* = \max\{0, -\frac{\mu + \lambda_k}{\kappa_k}\}$ where $\mu$ is found (by bisection) such that $\sum_{k=1}^K w_k^* = 1$
                \STATE Set $(b_1,\ldots, b_K) \gets \textrm{round}(b (w_1^*,\ldots, w_K^*))$
		        \STATE Sample $b_k$ data points from each task~$k$
                \STATE Compute the gradient $\bg$ using the estimates samples
                \STATE{Update: $\ttheta_{t+1} \gets \textrm{Optimizer} (\eta, \ttheta_t,\bg)$ } 
		\ENDFOR
    \end{algorithmic}
    \caption{Conceptual version of PiKE: Positive gradient Interaction-based K-task weights Estimator}
    \label{alg: Basic PiKE}
\end{algorithm}

As discussed in section~\ref{sec:PiKe}, this algorithm is computationally inefficient as it requires estimating $\nabla \cL_k(\theta_t)$ and $\sigma_k$ at each iteration. To improve efficiency, we introduced modifications that led to the development of the PiKE algorithm (Algorithm~\ref{alg: main} in the main body).

\section{Balanced-PiKE: Fairness Considerations Across Tasks}\label{app:pike-fairness}
Here, we present the \textit{Balanced-PiKE} algorithm in more detail. As discussed in the main body, the main difference with PiKE is that the fair version requires the computation of the coefficients
\[
y_k^\star =\frac{\tau e^{\tau\cL_k(\ttheta)-1}}{\sum^K_{k=1} e^{\tau\cL_k(\ttheta) -1}},\forall k
\]
Then updating the sampling weights by
\[
w_k \gets w_k \exp\left(
     (y_k^\star)^2 \zeta_1 \|\nabla \cL_k(\bw)\|^2 - (y_k^\star)^2\frac{\zeta_2}{2b} \sigma_k^2
    \right),\;\;\;\forall k
\]
The overall algorithm is summarized in Algorithm~\ref{alg: fair_main}.
For our experiments, we evaluate three different values of $\tau$: 1, 3, and 5. A larger $\tau$ results in a stronger balancing effect between different tasks.

\begin{algorithm}[H] 
    \begin{algorithmic}[1]
        \STATE {{\bfseries Input:} $\ttheta$, $T_0$, total batch size $b$, task $k$ dataset $\cD_k$, hyperparameters $\zeta_1\, \zeta_2, \tau$, prior weights~$\bw'$ } 
	\STATE {{\bfseries Initialize:} $w_k \gets 1/K$ or $w_k \gets w_k'$ } 
		\FOR{$t=0,1, \dots$}
                \IF{$t \mod T_0 = 0$}
                    \STATE Estimate $\|\nabla \cL_k(\ttheta_t)\|^2$, $\sigma_k^2$, and $\cL_k(\ttheta_t)$ for every $k$
                    \STATE $y_k^\star =\frac{\tau e^{\tau\cL_k(\ttheta)}}{\sum^K_{k=1} e^{\tau\cL_k(\ttheta) }}$
                    \STATE $w_k \gets w_k \exp\left(
                     (y_k^\star)^2 \zeta_1 \|\nabla \cL_k(\bw)\|^2 - (y_k^\star)^2\frac{\zeta_2}{2b} \sigma_k^2
                    \right)$
                    
                    \STATE  $\bw \gets \bw /\|\bw\|_1$
                    \STATE $(b_1,\ldots, b_K) \gets \textrm{round}(b (w_1,\ldots, w_K))$
                \ENDIF
		        \STATE Sample $b_k$ data points from each task~$k$
                \STATE Compute the gradient $\bg$ using the estimates samples
                \STATE{Update: $\ttheta_{t+1} \gets \textrm{Optimizer} (\eta, \ttheta_t,\bg)$} 
		\ENDFOR
    \end{algorithmic}
    \caption{\textit{Balanced-PiKE}: Balanced considerations across tasks}
    \label{alg: fair_main}
\end{algorithm}

\section{Experiments Setup}\label{app:experiment_setup}
\subsection{Dataset Details}
Our experiments construct two primary scenarios for multitask learning: multilingual tasks and diverse task mixtures spanning multiple domains. We consider two widely-used datasets for our study: mC4~\citep{xue2020mt5} and GLaM~\citep{du2022glam}.

\textbf{mC4 Dataset} The mC4 dataset~\citep{xue2020mt5} is a multilingual text corpus derived from the Common Crawl web archive, covering a diverse range of languages. It has been widely used for pretraining multilingual models, such as mT5~\citep{xue2020mt5} and ByT5~\citep{xue2021byt5}. The dataset is curated by applying language-specific filtering to extract high-quality text, ensuring a balanced representation across languages. Mixture weights for training models on mC4 are often chosen based on token counts. In our cases, we mainly focus on English (en), Hindi (hi), and German (de). We report their details in Table~\ref{tab: mc4-dataset-overview}. 

\begin{table*}[htb]
\centering
\small
\caption{Partial statistics of the mC4 corpus, totaling 6.3T tokens.} 
\begin{tabular}{lcc}
\toprule
ISO code & Language & Tokens (B)  \\
\midrule
en & English & 2,733 \\
hi & Hindi & 24 \\
de  & German & 347  \\
\bottomrule
\end{tabular}
\vskip 0.1in
\label{tab: mc4-dataset-overview}
\end{table*}

\textbf{GLaM Dataset} The GLaM dataset~\citep{du2022glam} comprises English text from six distinct sources and has been used to train the GLaM series models and PaLM~\citep{chowdhery2023palm}. Mixture weights for GLaM training were determined based on small model performance~\citep{du2022glam}, while \citep{xie2024doremi} employed group distributionally robust optimization (Group DRO) to compute domain-specific weights. Table~\ref{tab: glam-dataset-overview} summarizes the six domains in the GLaM dataset and the mixture weights selected by GLaM and DoReMi. We use these weights as oracle baselines for comparison with PiKE, which dynamically adjusts task weights over time using gradient information, unlike the fixed weights employed by GLaM and DoReMi.

\begin{table*}[htb]
\centering
\small
\caption{GLaM dataset~\citep{du2022glam} and fixed mixture weights used in GLaM~\citep{du2022glam} and DoReMi~\citep{xie2024doremi}.} 
\vskip 0.1in
\label{tab: glam-dataset-overview}
\begin{tabular}{lcccc}
\toprule
Dataset & Tokens (B) & Weight chosen by GLaM \citep{du2022glam} & Weight chosen by DoReMi \citep{xie2024doremi} \\
\midrule
Filtered Webpages & 143 & 0.42 & 0.51 \\
Wikipedia & 3 & 0.06 & 0.05\\
Conversations  & 174 & 0.28 & 0.22 \\
Forums & 247 & 0.02 & 0.04\\
Books & 390 & 0.20 & 0.20\\
News & 650 & 0.02  & 0.02\\
\bottomrule
\end{tabular}
\end{table*}

\subsection{Training Details}

Our experiments explore two distinct scenarios for multitask learning: multilingual training and diverse task mixtures spanning multiple domains. To achieve optimal results, we customize the training setups for each scenario and present them separately in this section. All training is performed from scratch.

\textbf{Multilingual Training} To address the complexities of tokenizing multilingual data, we utilize the mT5 tokenizer~\citep{xue2020mt5}, which features a vocabulary size of 250K. Both GPT-2 small and GPT-2 large models are trained with a context length of 1024 and a batch size of 256. The AdamW optimizer~\citep{loshchilov2018decoupled} is employed with consistent hyperparameters and a learning rate scheduler. Additional details on hyperparameter configurations are provided in Appendix~\ref{app:hyperparams}.

\textbf{GLaM Training} For GLaM training, we use the T5 tokenizer~\citep{raffel2020exploring}, implemented as a SentencePiece tokenizer trained on the C4 dataset with a vocabulary size of 32,000. Both GPT-2 small and GPT-2 large models are trained with a context length of 1024 and a batch size of 256. The AdamW optimizer~\citep{loshchilov2018decoupled} is used, and additional details on hyperparameters is in Appendix~\ref{app:hyperparams}.

\subsection{Model Architecture}

The detailed architecture is summarized in Table~\ref{tab:archictectures}. Our implementation utilizes pre-normalization~\citep{radford2019language} Transformers with qk-layernorm~\citep{dehghani2023scaling}. Consistent with \cite{chowdhery2022palm}, we omit biases, and the layernorm~\citep{ba2016layer}  value remains set to the Flax~\citep{flax2020github} default of 1e-6. Additionally, we incorporate rotary positional embeddings~\citep{su2021roformer}.

\begin{table}
\caption{Architecture hyperparameters for different model scales used in the paper. All models are GPT-2-like decoder-only architectures. The multilingual models employ a vocabulary size of 250K, whereas GLaM training uses a vocabulary size of 32K. Differences in the total number of parameters arise due to the variation in vocabulary sizes.}
\label{tab:archictectures}
\centering
\vspace{0.2cm}
\begin{adjustbox}{max width=0.9\textwidth}
\begin{tabular}{lrrrrr}
\toprule
   Size &  \# Params & Layers & Attention heads & Attention head dim & Hidden dim \\
     \midrule
GPT-2 small & 110M/270M  & 12      & 12               & 64                      & 768       \\
GPT-2 large &  750M/1B & 36      & 20               & 64                & 1280       \\
\bottomrule
\end{tabular}
\end{adjustbox}
\end{table}

\subsection{Experimental Resource}
All experiments are conducted on 8 Google TPUv4. The training time for GPT-2 small and GPT-2 large models for 120K steps are approximately 1 day and 2 days per run, respectively.

\subsection{Hyper-parameters}\label{app:hyperparams}
Table~\ref{tab:hyperparameters} shows the detailed hyperparameters that we used in all our experiments. We also report our hyperparameters grid for tuning PiKE in Table~\ref{tab:hyperparameters_pike}. 
\begin{table}[!tb]
    \caption{Hyperparameter settings for our experiments.}
    \vspace{0.2cm}
    \centering
    \small
    \begin{tabular}{ll}
    \toprule
    \textbf{Hyperparameters} & \textbf{Values} \\
    \midrule
    Optimizer & AdamW ($\beta_1=0.95$, $\beta_2=0.98$) \\
    Initial and final learning rate & $7e-6$ \\
    Peak learning rate & $7e-4$ \\
    Weight decay & $0.1$ \\
    Batch size & $256$ \\ 
    Context length & $1024$ \\
    Gradient clipping norm & $1.0$ \\
    Training step & $120,000$ \\
    Warm-up step & $10,000$\\
    Schedule & Linear decay to final learning rate \\
    \bottomrule
    \end{tabular}
    \label{tab:hyperparameters}
\end{table}

\begin{table}[!tb]
    \caption{Hyperparameter settings for running PiKE (Algorithm~\ref{alg: main}).}
    \vspace{0.2cm}
    \centering
    \small
    \begin{tabular}{ll}
    \toprule
    \textbf{Hyperparameters} & \textbf{Values} \\
    \midrule
    PiKE hyperparameter $(\zeta_1, \zeta_2)$ & $\{(1e-2, 1e-3), (1.5e-2, 1e-3), (5e-2, 5e-3),$  \\ & $\ (7.5e-2,5e-3), (1e-1, 1e-2), (1.5e-1, 1e-2)\}$ \\
    Check interval $T_0$ & 1,000 \\
    Batch size for estimation & 256
    \\ \bottomrule
    \end{tabular}
    \label{tab:hyperparameters_pike}
\end{table}

\subsection{Implementation Details}

Our implementation builds upon the Nanodo training infrastructure~\citep{wortsman2023small}, incorporating enhancements for efficiency. This framework relies on Flax~\citep{flax2020github}, JAX~\citep{jax2018github}, and TPUs~\citep{jouppi2017datacenter}.

To enable training of larger models, we shard both model and optimizer states, following the methodology of FSDP~\citep{ren2021zero}, and define these shardings during JIT compilation. Checkpointing is handled using Orbax~\citep{orbax}, while deterministic data loading is facilitated by Grain~\citep{grain}.

For data loading, sequences are packed to avoid padding. When a sequence contains fewer tokens than the context length hyperparameter, an end-of-sequence token is appended. This differs from Nanodo~\citep{wortsman2023small}, where both begin-of-sequence and end-of-sequence tokens are added.

\subsection{Evaluation}

Our evaluation adheres to the OLMES suite~\citep{gu2024olmes}. For multilingual downstream performance, we utilize the multilingual version of HellaSwag~\citep{dac2023okapi}, which supports evaluations across 26 languages. English downstream tasks are assessed using ARC-Easy~\citep{clark2018think}, CommonsenseQA~\citep{talmor2018commonsenseqa}, PIQA~\citep{bisk2019reasoning}, and HellaSwag~\citep{zellers2019hellaswag}. Unless specified otherwise, multilingual evaluations are performed in a 0-shot setting, while GLaM pretraining evaluations employ 7-shot in-context learning, with demonstration candidates separated by two line breaks. For HellaSwag and its translated variants, we evaluate the first 3,000 examples. For all other downstream tasks, evaluations are conducted on their respective validation sets. In the case of multiple-choice tasks, different candidates are included in the prompt, and the average log-likelihood for each candidate is computed. The candidate with the highest score is then selected as the predicted answer.

\section{Additional Experiment Results}

\subsection{Comparison of Performance Using Mix, Random, and Round-Robin Sampling Strategies}\label{app: mix_rr_random}
\begin{figure*}[!htb]
\centering
\begin{minipage}{.5\textwidth}
  \centering
  \includegraphics[width=0.8\linewidth]{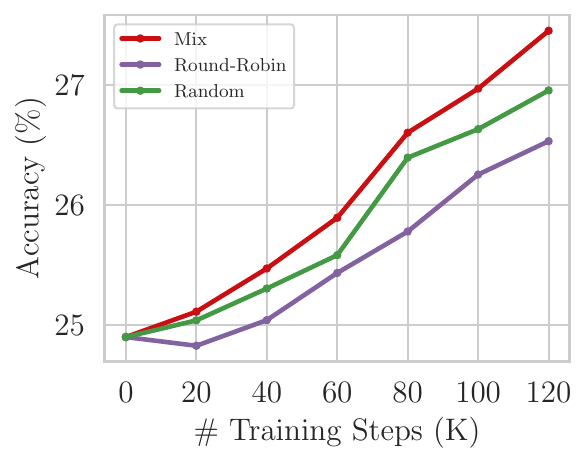}
  \captionof{subfigure}{1B models on multilingual C4 (en), C4 (hi), and C4 (de) datasets }
\end{minipage}%
\begin{minipage}{.5\textwidth}
  \centering
  \includegraphics[width=0.8\linewidth]{figures/glam_large_convergence_smooth.pdf}
  \captionof{subfigure} {750M models on GLaM datasets with six domains}
\end{minipage}
\caption{Average downstream task accuracy of pretraining language models using Mix, Round-Robin, and Random sampling strategies. Mix and Random use equal batch size for each task ($b_k = b/K, \forall k \in K$).}
\label{fig:app_mix_round_rr_comparison}
\end{figure*}

Figure~\ref{fig:app_mix_round_rr_comparison} presents the average downstream accuracies of language models pre-trained using Mix, Random, and Round-Robin sampling strategies. In both multilingual pre-training and GLaM pre-training, the Mix sampling strategy consistently outperforms the other two. This motivates us its use in pre-training large language models.

\subsection{Cosine Similarity and $\ulc$-Conflicted Gradients} \label{app: cos_sim_grad_conflict}
\begin{figure*}[!htb]
\centering
\begin{minipage}{.5\textwidth}
  \centering
  \includegraphics[width=0.8\linewidth]{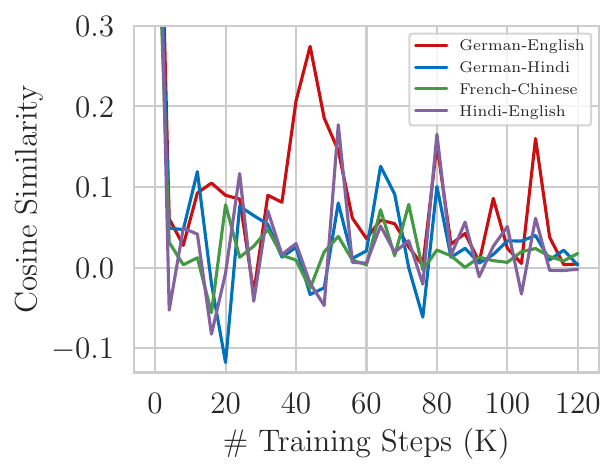}
\end{minipage}%
\begin{minipage}{.5\textwidth}
  \centering
  \includegraphics[width=0.8\linewidth]{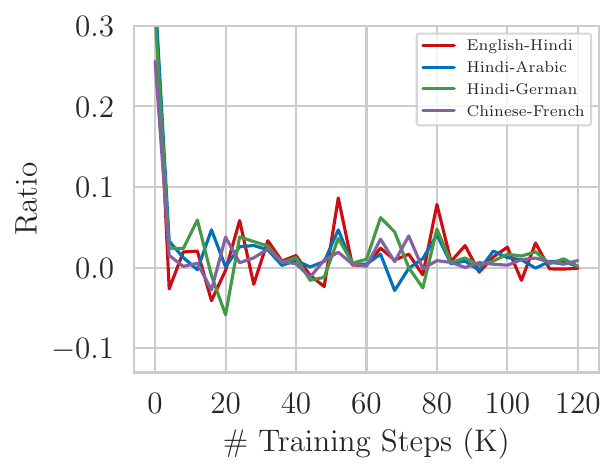}
\end{minipage}
\caption{1B models trained on multilingual mC4 datasets. \textbf{Left:} Cosine similarity between task gradients during language model pre-training over time. \textbf{Right:} The ``ratio,'' which defined as  $
\frac{\lin{\cL_j(\ttheta),\cL_k(\ttheta)}}{\|\cL_j(\ttheta)\|^2 +\| \cL_k(\ttheta)\|^2},$ between task gradients during language model pre-training over time. ``\textit{data1-data2}'' denotes the cosine similarity or ratio between the gradient of \textit{data1} and the gradient of \textit{data2}.}

\label{fig:app_mutlilingual_grad_similarity}
\end{figure*}

\begin{figure*}[!htb]
\centering
\begin{minipage}{.5\textwidth}
  \centering
  \includegraphics[width=0.8\linewidth]{figures/glam_cos_similarity.pdf}
\end{minipage}%
\begin{minipage}{.5\textwidth}
  \centering
  \includegraphics[width=0.8\linewidth]{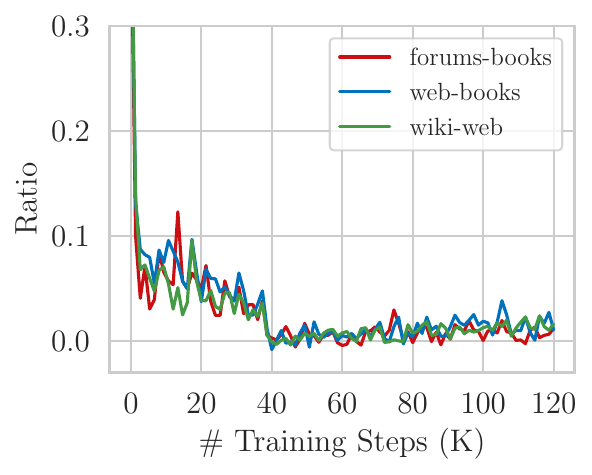}
\end{minipage}
\caption{750M models on GLaM datasets with six domains. \textbf{Left:} Cosine similarity between task gradients during language model pre-training over time. \textbf{Right:} The ``ratio,'' which defined as  $
\frac{\lin{\cL_j(\ttheta),\cL_k(\ttheta)}}{\|\cL_j(\ttheta)\|^2 +\| \cL_k(\ttheta)\|^2},$ between task gradients during language model pre-training over time. ``\textit{data1-data2}'' denotes the cosine similarity or ratio between the gradient of \textit{data1} and the gradient of \textit{data2}.}
\label{fig:app_glam_grad_similarity}
\end{figure*}

Figures~\ref{fig:app_mutlilingual_grad_similarity} and~\ref{fig:app_glam_grad_similarity} show the cosine similarity, defined as $\frac{\lin{\cL_j(\ttheta),\cL_k(\ttheta)}}{\|\cL_j(\ttheta)\|\|\cL_k(\ttheta)\|} $ 
and the ``ratio,'' defined as  
$\frac{\lin{\cL_j(\ttheta),\cL_k(\ttheta)}}{\|\cL_j(\ttheta)\|^2 +\| \cL_k(\ttheta)\|^2}. $ In particular, if  
$\frac{\lin{\nabla \cL_j(\ttheta) ,\nabla \cL_k(\ttheta)}}{\|\cL_j(\ttheta) \|\|\cL_k(\ttheta) \|} \geq -\tilde{c},$ then the gradients are $\ulc$-conflicted for $\ulc = \tilde{c}/2$, which aligns with the observations in Figures~\ref{fig:app_mutlilingual_grad_similarity} and~\ref{fig:app_glam_grad_similarity}.

\subsection{Comparison of Performance Using PCGrad, AdaTask, and Mix}

\begin{figure}[!htb]
\vskip 0.2in
\begin{center}
\centerline{\includegraphics[width=0.4\columnwidth]{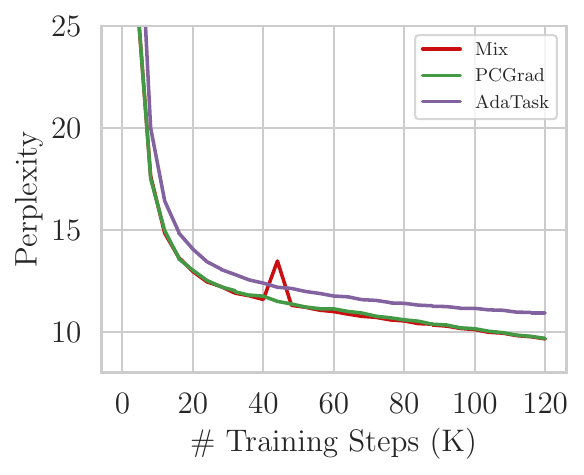}} 
\caption{Eval perplexity of pretraining 270M GPT-2 style multilingual language models on mC4 datasets (English and Hindi) using Mix, PCGrad, and AdaTask.}
\label{fig: motivation_different_mtl}
\end{center}
\end{figure}

Figure~\ref{fig: motivation_different_mtl} presents the average downstream task performance on HellaSwag (en) and HellaSwag (hi) for 270M multilingual language models pre-trained using PCGrad, AdaTask, and Mix. As shown in Figure~\ref{fig: motivation_different_mtl}: 1) PCGrad performs similarly to Mix, as it only adjusts gradients when conflicts occur—which is rare. 2) AdaTask converges more slowly due to noisy gradients and suboptimal optimizer state updates. Additionally, both methods are memory-intensive, requiring \(O(K)\) storage for task gradients (PCGrad) or optimizer states (AdaTask), making them impractical for large-scale models such as the 540B PaLM~\citep{chowdhery2022palm}.

\subsection{PiKE: Minimal Training and Memory Overhead} \label{app:pike-training-overhead}
Table~\ref{tab:trainig-overhead} reports the total training time and the computational overhead incurred by PiKE. In our experiments, per-task gradient squared norms and variances are estimated only once every $T_0 = 1,000$ training steps. Profiling results indicate that this infrequent estimation introduces minimal overhead during the pre-training of LLMs. Specifically, for a 1B parameter multilingual model pre-trained for approximately 71 hours, this estimation process accounted for only about 2.3\% of the total training time (roughly 1.6 hours). This overhead percentage can be further reduced by using a larger estimation interval, $T_0$ (e.g., 5,000 steps), or by adopting a more efficient parallelization implementation for handling multiple tasks.

Regarding memory requirements, PiKE only needs to store the periodically estimated scalar values for each task's gradient variance and squared norm. This amounts to an additional memory footprint of $\mathcal{O}(K)$, which is negligible compared to the model parameters ($\mathcal{O}(d)$) and especially to the $\mathcal{O}(Kd)$ memory demanded by prior MTL methods that store $K$ full per-task gradients. This efficiency starkly contrasts with those earlier approaches, which also often involve solving auxiliary optimization problems that further increase computational and memory burdens~\citep{xin2022current}. Unlike approaches such as DoReMi or GLaM, PiKE simplifies deployment due to its automated learning of task sampling weights during the training procedure. This dynamic weight adaptation facilitates straightforward implementation and seamless integration into existing workflows, eliminating the need for manual intervention or training disruptions.

\begin{table*}[h]
\centering
\caption{We report the total training time (hrs) and how much overhead time that running PiKE (hrs). Compared with the total training, the overhead for running PiKE is minimal only taking from 1.2\% to 2.4\% in training large models.} \label{tab:trainig-overhead}
{\small
\begin{adjustbox}{max width=\textwidth}
\begin{tabular}{@{}lccc@{}}
\toprule
                   & Total Training Time (hrs) & Overhead time for running PiKE (hrs) & Overhead time for running PiKE (\%) \\ \midrule
GPT-2 small (110M, GLaM) & 10                        & 0.12        & 1.2                         \\ 
GPT-2 small (270M, Multilingual) & 12                        & 0.24     & 2                            \\
GPT-2 large (750M, GLaM) & 41                        & 1.0  & 2.4                                \\
GPT-2 large (1B, Multilingual)   & 71                        & 1.6      & 2.3                         \\ \bottomrule
\end{tabular}
\end{adjustbox}
}
\end{table*}

\subsection{Additional Zero-shot and Seven-shot Results} \label{pike-zero-seven-shot-results}

In the main text, our evaluation focused on 0-shot performance for multilingual pre-training and 7-shot performance for GLaM pre-training. This appendix provides a more complete picture by presenting both 0-shot and 7-shot evaluation results for the checkpoints from these respective experimental setups. The results are in the Table~\ref{tab:0-7-shots-multilingual} and~\ref{tab:0-7-shots-glam}. 

\begin{table*}[!tb]
\large
\centering
\caption{We compare the accuracies (\%, higher the better) of different models on HellaSwag and its corresponding translated version using 0- and 7-Shot settings.~\textbf{Bolding} indicates the best model in the task; $\ols{\text{Metrics}}$ means the average across different tasks.}
\label{tab:0-7-shots-multilingual}
\begin{adjustbox}{max width=\textwidth}
\begin{tabular}{lcccc|cccc}
\toprule
            &               & HellaSwag (en)       & HellaSwag (hi)       & HellaSwag (de) &               & HellaSwag (en) & HellaSwag (hi) & HellaSwag (de) \\ \cmidrule(lr){3-5} \cmidrule(l){7-9} 
 &
  $\ols{\text{Accuracy} \uparrow}$ &
  0-shot $\uparrow$ &
  0-shot $\uparrow$ &
  0-shot $\uparrow$ &
  $\ols{\text{Accuracy}(\%) \uparrow}$ &
  7-shot $\uparrow$ &
  7-shot  $\uparrow$ &
  7-shot $\uparrow$ \\ \midrule
\multicolumn{9}{l}{\textbf{C4 (en), C4 (hi), and C4 (de) datasets, GPT-2 large style, 1B params, 36 Layers default, 120K training steps}}                     \\
Mix & 27.5 & 28.1 & 27.1 & \textbf{27.6} & 32.6 & 33.9 & \textbf{31.4} & \textbf{32.5} \\
Round-Robin & 26.5 & 27.6 & 26.7 & 26.3 & 32.0 & 34.0 & 29.7 & 32.4 \\
Random & 26.6 & 27.0 & 26.9 & 26.1 & 31.9 & 34.0 & 30.0 & 31.7 \\
PiKE  & \textbf{28.7} & \textbf{33.0} & \textbf{27.2} & 26.2 & \textbf{33.2} & \textbf{39.0} & 29.6 & 30.9 \\
\bottomrule
\end{tabular}
\end{adjustbox}
\label{tab:partial_main-table-multilingual-0-7-shots}
\end{table*}

\begin{table*}[!tb]
\centering

\caption{We compare the accuracies (\%, higher the better) of different models on four different Q/A tasks using 0- and 7-Shot settings. \textbf{Bolding} indicates the best model in the task, $\ols{\text{Metrics}}$ means the average across different tasks.}
\label{tab:0-7-shots-glam}
{\small
\begin{adjustbox}{max width=\textwidth}
\begin{tabular}{lccccc|ccccc}
\toprule
            &               & ArcE       & CSQA       & HellaSwag & PIQA &              & ArcE       & CSQA       & HellaSwag & PIQA \\ 
            \cmidrule(l){3-6} \cmidrule(l){8-11} 
 &
  $\ols{\text{Accuracy} \uparrow}$ &
  0-shot $\uparrow$ &
  0-shot $\uparrow$ &
  0-shot $\uparrow$ &
  0-shot $\uparrow$ &
  $\ols{\text{Accuracy}(\%) \uparrow}$ &
  7-shot $\uparrow$ &
  7-shot  $\uparrow$ &
  7-shot  $\uparrow$ &
  7-shot $\uparrow$ \\ \midrule
\multicolumn{9}{l}{\textbf{Six domains of GLaM dataset, GPT-2 large style, 750M params, 36 layers default}}                     \\
Mix & 33.6 & 30.3 & 20.8 & 29.5 & 53.8 & 46.4 & 47.2 & 39.6 & 37.9 & 60.9 \\
Round-Robin & 32.7 & 30.4 & 20.3 & 26.1 & 53.9 & 44.3 & 43.5 & 36.7 & 36.8 & 60.3 \\
Random & 32.0 & 28.9 & 20.5 & 26.2 & 52.3 & 42.7 & 41.7 & 34.2 & 36.6 & 58.2 \\
GLaM & 31.7 & 28.8 & 19.9 & 26.3 & 51.9 & 45.3 & 46.9 & 39.8 & \textbf{38.0} & 56.4 \\
DoReMi & 35.6 & 33.0 & 23.8 & 30.0 & 55.7 & 46.5 & 48.6 & 40.1 & 37.5 & 59.6 \\
PiKE (Uniform) & \textbf{37.9} & \textbf{37.4} & 24.2 & \textbf{33.9} & 56.1 & 47.6 & 49.6 & 43.2 & 37.2 & 60.4 \\
PiKE (GLaM) & 35.5 & 33.5 & 20.4 & 31.2 & \textbf{56.8} & \textbf{48.1} & \textbf{49.8} & \textbf{43.5} & \textbf{38.0} & \textbf{61.2} \\
\bottomrule
\end{tabular}
\end{adjustbox}
}
\label{tab:partial_main-table-glam-0-7-shots}
\end{table*}

\subsection{Ablation: Importance of Gradient Variance for PiKE} \label{app:importance-of-second-term}
This section underscores the importance of PiKE's per-task gradient variance component for its overall effectiveness. To demonstrate this, we conducted an ablation study where PiKE's $\zeta_1$ hyperparameter was held constant at $0.1$, while $\zeta_2$, which controls the influence of this gradient variance term, was varied. The specific case of $\zeta_2=0$ signifies the complete omission of the variance component from PiKE's formulation. The results are in Table~\ref{tab:importance_of_gradient_variance}. For instance, PiKE achieved a mean accuracy of $47.0\%$ with $\zeta_2=0.005$, which dropped to $45.3\%$ when the variance term was omitted ($\zeta_2=0$). This performance degradation highlights the critical role of per-task gradient variance in regulating PiKE's sampling weights, affirming its necessity for achieving optimal results.

\begin{table}[!tb]
\centering
\caption{Mean Accuracies (\%, higher values indicate better performance) of GLaM 740M models pre-trained with PiKE on four Question/Answering (Q/A) tasks, under different PiKE hyperparameter settings. PiKE's $\zeta_1$ hyperparameter was fixed constant at $0.1$, while $\zeta_2$ was varied. A value of $\zeta_2=0$ means an ablation where the per-task gradient variance term in PiKE is omitted.}
\label{tab:importance_of_gradient_variance}

\begin{tabular}{@{}ccccccc@{}}
\toprule
$\zeta_2$ & 0    & 0.001 & 0.005 & 0.01 & 0.05 & 0.1  \\ \midrule
\text{Accuracy} & 45.3 & 46.2  & \textbf{47.0}  & 46.4 & 46.9 & 45.3 \\ \bottomrule
\end{tabular}
\end{table}

\subsection{Details on Estimating Gradient Variance and Magnitude, and Sensitivity Analysis} \label{app:pike-variance-magnitude-estimation-details}

We estimate the per-task gradient variance and magnitude (measured using the $L_2$ norm) every $T_0 = 1000$ training steps. At each estimation step, the number of samples used for each task corresponds to its assigned portion ($b_k$) of the standard mixed training batch. Our JAX-based implementation enables efficient computation of these statistics with negligible overhead relative to total training time.

To evaluate the impact of estimation accuracy on PiKE's performance, we conduct a sensitivity analysis while keeping all other experimental settings and hyperparameters fixed. Specifically, we compare two strategies. The first, used in our main experiments, computes statistics for each task once using samples from a single training batch. The second estimates statistics multiple times using different subsets of samples (e.g., distinct micro-batches) and averages them to produce the final values used by PiKE. Results are shown in Table~\ref{tab:estimation_sensitivity_analysis}. We find that PiKE is robust to the estimation noise: computing gradient statistics from a single batch is sufficient to achieve strong performance, making the method both effective and efficient.

\begin{table}[!tb]
\centering
\caption{Mean accuracies (\%, higher is better) of GLaM 740M models pretrained with PiKE on four question answering (Q/A) tasks, evaluated under different PiKE hyperparameter settings. The table compares the effect of estimating gradient statistics for each task either once (using a single batch) or multiple times using different subsets of samples.}
\label{tab:estimation_sensitivity_analysis}

\begin{tabular}{@{}ccccc@{}}
\toprule
Number of Estimation & 1   & 2 & 3 & 4  \\ \midrule
\text{Accuracy} & 46.8 & 47.4  & 47.1  & 46.8 \\ \bottomrule
\end{tabular}
\end{table}

\subsection{Full Pre-training Results}
Tables~\ref{tab:main-table-multilingual} and~\ref{tab:main-table-glam} present the complete results of pre-training language models across various scales (110M, 270M, 750M, and 1B) and scenarios (Multilingual and GLaM datasets). PiKE consistently outperforms all baselines across all scales and scenarios.

\begin{table*}[!tb]
\large
\centering
\caption{We report the perplexities (lower the better) on the validation split of multilingual C4 datasets. We also compare the accuracies (\%, higher the better) of different models on HellaSwag and its corresponding translated version. HellaSwag and its translated versions have 4 choices. \textbf{Bolding} indicates the best model in the task, $\ols{\text{Metrics}}$ means the average across different tasks.}
\begin{adjustbox}{max width=\textwidth}
\begin{tabular}{lcccc|cccc}
\toprule
            &               & C4 (en)       & C4 (hi)       & C4 (de) &               & HellaSwag (en) & HellaSwag (hi) & HellaSwag (de) \\ \cmidrule(lr){3-5} \cmidrule(l){7-9} 
 &
  $\ols{\text{Perplexity} \downarrow}$ &
  Perplexity $\downarrow$ &
  Perplexity $\downarrow$ &
  Perplexity $\downarrow$ &
  $\ols{\text{Accuracy}(\%) \uparrow}$ &
  0-shot $\uparrow$ &
  0-shot  $\uparrow$ &
  0-shot $\uparrow$ \\ \midrule
\multicolumn{9}{l}{\textbf{Single dataset, GPT-2 small style, 270M params, 12 layers default, 120K training steps}}                                           \\
C4 (en)     &  13.25              &     13.25          & *             & *       &       26.5        &      26.5          & *              & *              \\
C4 (hi)     &   4.97            & *             &     4.97          & *       &     26.4          & *              &     26.4           & *              \\
C4 (de)     &    11.27           & *             & *             &  11.27        &  26.1             & *              & *              &     26.1           \\ \midrule
\multicolumn{9}{l}{\textbf{C4 (en) and C4 (hi) datasets, GPT-2 small style, 270M params, 12 layers default, 120K training steps}}                             \\
Mix         & 10.50         & 15.46         & \textbf{5.55}          & *       & 25.5          & 24.4           & 26.5           & *              \\
Round-Robin & 10.57         & 15.57         & 5.57          & *       & 25.6          & 25.2           & 26.0           & *              \\
Random      & 10.57         & 15.57         & 5.57          & *       & 25.3          & 24.3           & 26.3           & *              \\
FAMO~\cite{liu2024famo}  & 10.38 & 15.18 & 5.57 & * & 25.7   & 24.8 & 26.5 & * \\
ADO~\cite{jiang2024adaptive} & 10.45 & 15.39 & 5.52 & * & 25.1 & 24.3 &  25.8 & * \\
PiKE       & \textbf{10.15}         & \textbf{14.31}         & 5.99          & *       & \textbf{26.5}          & \textbf{26.0}           & \textbf{27.0}           & *              \\ \midrule
\multicolumn{9}{l}{\textbf{C4 (en), C4 (hi), and C4 (de) datasets, GPT-2 small style, 270M params, 12 layers default, 120K training steps}}                   \\
Mix         &  12.00             & 16.30               &\textbf{5.88}               & 13.83         &25.3               &24.4                &26.0                &\textbf{25.5}                \\
Round-Robin &12.10               &16.44               &5.91               &13.95         &25.1               &24.3                &26.0                &24.9                \\
Random      &12.16               &16.49               &5.95               &14.03         &25.1               &24.7                &\textbf{26.6}                &23.9                \\
FAMO~\cite{liu2024famo}  & \textbf{11.92}  & 16.19 & 6.00 & \textbf{13.57} & 24.8 & 24.5 & 25.2 & 24.8 \\
ADO~\cite{jiang2024adaptive} & 12.01 & 16.31 & \textbf{5.88} & 13.84 & 24.9 & 24.4 & 25.4 & 24.8 \\
PiKE       &12.01               &\textbf{15.48}               &5.92               &14.64         &\textbf{25.6}               &\textbf{25.4}                &26.4                &24.8                \\ \midrule
\multicolumn{9}{l}{\textbf{Single dataset, GPT-2 large style, 1B params, 36 Layers default, 120K training steps}}                                             \\
C4 (en)     & 9.30          & 9.30          & *             & *       & 33.6          & 33.6           & *              & *              \\
C4 (hi)     & 3.87          & *             & 3.87          & *       & 27.5          & *              & 27.5           & *              \\
C4 (de)     &  7.72         & *             & *            &  7.72   &  28.1         & *              & *              &      28.1      \\ \midrule
\multicolumn{9}{l}{\textbf{C4 (en) and C4 (hi) datasets, GPT-2 large style, 1B params,  36 Layers default, 120K training steps}}                              \\
Mix         & 7.41          & 10.60         & \textbf{4.22} & *       & 27.3          & 28.2           & 26.5           & *              \\
Round-Robin & 7.49          & 10.72         & 4.25          & *       & 27.5          & 28.0           & 27.0           & *              \\
Random      & 7.52          & 10.76         & 4.28          & *       & 28.0          & 28.9           & 27.0           & *              \\
FAMO~\cite{liu2024famo}  & 7.33 & 10.44 & \textbf{4.22} & * & 26.8 & 27.1 & 26.5 & * \\
ADO~\cite{jiang2024adaptive} & 7.41 & 10.59 & 4.23  & * & 26.5 & 26.0 & 26.9 & * \\
PiKE       & \textbf{7.21} & \textbf{9.63} & 4.80          & *       & \textbf{30.0} & \textbf{32.7}  & \textbf{27.3}  & *              \\ \midrule
\multicolumn{9}{l}{\textbf{C4 (en), C4 (hi), and C4 (de) datasets, GPT-2 large style, 1B params, 36 Layers default, 120K training steps}}                     \\
Mix & 8.29 & 11.13 & \textbf{4.45} & 9.29 & 27.5 & 28.1 & 27.1 & 27.6 \\
Round-Robin & 8.41 & 11.31 & 4.97 & 9.46 & 26.5 & 27.6 & 26.7 & 26.3 \\
Random & 8.48 & 11.38 & 4.54 & 9.55 & 26.6 & 27.0 & 26.9 & 26.1 \\
FAMO~\cite{liu2024famo}  & 8.25 & 11.04 & 4.48 & 9.23 & 27.2 & 27.3 & 26.9 & 27.3 \\
ADO~\cite{jiang2024adaptive} & 8.30 & 11.12 & 4.45 & 9.31 & 27.5 & 27.7 & \textbf{27.5} & 27.2  \\
PiKE & 9.56 & \textbf{9.49} & 5.32 & 13.87 & 28.7 & \textbf{33.0} & 27.2 & 26.2 \\ 
Balanced-PiKE ($\tau=1$) & 8.29 & 11.12 & 4.46 & 9.31 & 27.9 & 28.3 & 27.4 & 28.0 \\
Balanced-PiKE ($\tau=3$) & \textbf{8.18} & 10.14 & 4.93 & 9.49 & \textbf{28.9} & 31.3 & 27.3 & 28.1 \\
Balanced-PiKE ($\tau=5$) & 8.42 & 10.02 & 6.30 & \textbf{8.94} & \textbf{28.9} & 31.2 & 26.9 & \textbf{28.6} \\
\bottomrule
\end{tabular}
\end{adjustbox}
\label{tab:main-table-multilingual}
\end{table*}

\begin{table*}[!tb]
\centering

\caption{We report perplexity (lower is better) on the validation split of the GLaM datasets, averaging perplexities across six domains when applicable or reporting a single perplexity when only training with a single domain. We also compare the accuracies (\%, higher the better) of different models on four different Q/A tasks. HellaSwag and ArcE tasks have 4 choices, CSQA has 5 choices, and PIQA has 2 choices. PiKE (Uniform) means PiKE using initial sampling weights of $1/6$ for each task and PiKE (GLaM) means PiKE using GLaM tuned weights as initial task weights. \textbf{Bolding} indicates the best model in the task, $\ols{\text{Metrics}}$ means the average across different tasks, \ul{underlining} indicates PiKE beating Mix, Round-Robin, Random methods}
\begin{adjustbox}{max width=\textwidth}
\begin{tabular}{lc|ccccc}
\toprule
                 & GLaM              &                      & ArcE                 & CSQA            & HellaSwag                 & PIQA                 \\ \cmidrule(l){4-7} 
 &
  $\ols{\text{Perplexity} \downarrow } $ &
  $\ols{\text{Accuracy} (\%) \uparrow}$ &
  7-shot $\uparrow$ &
  7-shot $\uparrow$ &
  7-shot  $\uparrow$ &
  7-shot $\uparrow$ \\ \midrule
\multicolumn{7}{l}{\textbf{Single domain of GLaM dataset, GPT-2 small style, 110M params, 12 layers default }}                         \\
Wikipedia           &  9.96 & 33.5 & 32.5 & 20.9 & 27.3 & 53.3 \\
Filtered Webpage    & 16.05 & 37.2 & 38.4 & 26.8 & 27.6 & 55.8 \\
News                & 9.33 & 33.8 & 31.1 & 22.7 & 27.0 & 54.5 \\
Forums              & 22.87 & 35.5 & 32.1 & 23.4 & 28.7 & 57.6 \\ 
Books               & 16.81 & 34.7 & 34.3 & 22.1 & 27.8 & 54.7 \\
Conversations       & 18.27 & 36.1 & 32.6 & 25.6 & 28.6 & 57.6 \\ \midrule
\multicolumn{7}{l}{\textbf{Six domains of GLaM dataset, GPT-2 small style, 110M params, 12 layers default}}                       \\
Mix&  18.27                   &36.2                   &35.6                       &24.1                      &\textbf{28.5}                      &56.7                      \\
Round-Robin      &  18.45                     &35.9                      &35.8                      &24.2                             &27.5                      &56.0                      \\
Random           &  18.48                    &35.5                      &34.3                      &22.4                      &28.4                      &56.8                      \\
FAMO~\cite{liu2024famo} & \textbf{18.19}  & 35.9  & 35.3  & 24.2   &  27.8 & 56.4         \\
ADO~\cite{jiang2024adaptive} & 18.27  & 36.2 & 35.7  & 24.8   & 27.7 & 56.4         \\

GLaM~\cite{du2022glam}             &   18.91                   &35.8                      &35.3                      &24.1                      &28.5                      &55.1\\
DoReMi~\cite{xie2024doremi}           &   18.98                   &37.0                     &36.0                      &\textbf{28.3}                     &28.2                      &55.3\\
PiKE (Uniform)             & 18.44           &\ul{37.4}                      &\ul{36.8}                      &\ul{27.5}                     &\ul{\textbf{28.5}}                      &\ul{\textbf{57.0}}   \\
PiKE (GLaM)             &    19.34           &\ul{\textbf{37.8}}                      &\ul{\textbf{39.0}}                      & \ul{27.0}                      &28.0                      &\ul{\textbf{57.0}}                      \\ \midrule
\multicolumn{7}{l}{\textbf{Single domain of GLaM dataset, GPT-2 large style, 750M params, 36 layers default}}                         \\
Wikipedia & 7.24 & 35.9 & 35.1 & 24.0 & 30.5 & 53.9       \\
Filtered Webpage &  11.12 & 40.9 & 36.7 & 33.2 & 34.2 & 56.5 \\
News  & 6.62 & 37.4 & 33.6 & 24.7 & 34.1 & 57.3 \\
Forums & 16.29 & 43.6 & 38.0 & 35.8 & 39.7 & 60.7          \\
Books & 11.83  & 41.3 & 40.0 & 33.0 & 34.5 & 57.8       \\
Conversations & 13.50 & 42.2 & 36.9 & 33.2 & 39.2 & 59.6     \\ \midrule
\multicolumn{7}{l}{\textbf{Six domains of GLaM dataset, GPT-2 large style, 750M params, 36 layers default}}                       \\
Mix &\textbf{12.77} & 46.4 & 47.2 & 39.6 & 37.9 & 60.9        \\
Round-Robin &12.98 & 44.3 & 43.5 & 36.7 & 36.8 & 60.3 \\
Random &12.99 & 42.7 & 41.7 & 34.2 & 36.6 & 58.2                 \\
FAMO~\cite{liu2024famo} & 13.25 & 45.0 & 43.7 & 40.0  & 36.4 & 59.8         \\
ADO~\cite{jiang2024adaptive} & \textbf{12.77} & 45.9 & 45.5 & 38.7   & 38.1  & 61.1           \\
GLaM~\cite{du2022glam} &13.20 & 45.3 &46.9 & 39.8 & 38.0 & 56.4            \\
DoReMi~\cite{xie2024doremi} &13.25 & 46.5 & 48.6 & 40.1 & 37.5 & 59.6         \\
PiKE (Uniform) &13.22 & \ul{47.6} &\ul{49.6} & \ul{43.2} & 37.2 &60.4   \\
PiKE (GLaM) &13.35 & \ul{48.1} & \ul{\textbf{49.8}} & \ul{\textbf{43.5}} & \ul{38.0} & \ul{61.2} \\
Balanced-PiKE ($\tau=1$) & 13.21 & \ul{47.5} & \ul{48.8} & \ul{42.5} & 37.6 & \ul{61.2} \\
Balanced-PiKE ($\tau=3$) & 13.26  & \ul{47.2} & \ul{48.8}  & \ul{41.5}   & 37.2 & \ul{61.3}  \\
Balanced-PiKE ($\tau=5$) & 13.19 & \ul{\textbf{48.2}} & \ul{49.3} & \ul{42.6} & \ul{\textbf{38.5}} & \ul{\textbf{62.4}} \\

\bottomrule
\end{tabular}
\end{adjustbox}
\label{tab:main-table-glam}
\end{table*}

\subsection{Adaptive Sampling Weights of PiKE During Pre-training}
\begin{figure*}[!htb]
\centering
\begin{minipage}{.33\textwidth}
  \centering
  \includegraphics[width=1\linewidth]{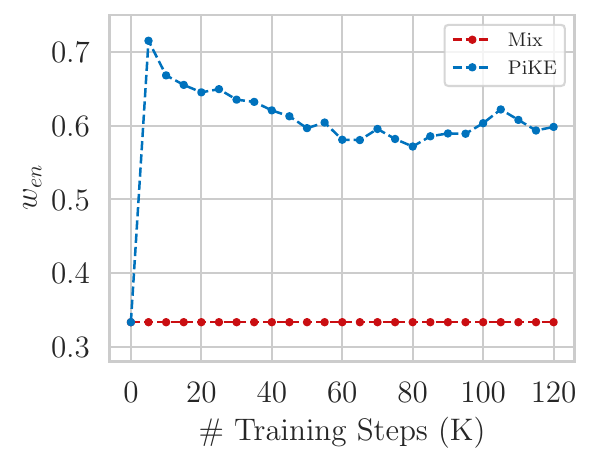}
\end{minipage}%
\begin{minipage}{.33\textwidth}
  \centering
  \includegraphics[width=1\linewidth]{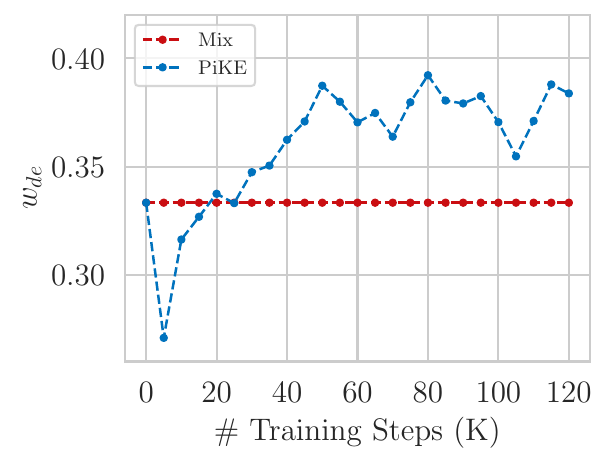}
\end{minipage}
\begin{minipage}{.33\textwidth}
  \centering
  \includegraphics[width=1\linewidth]{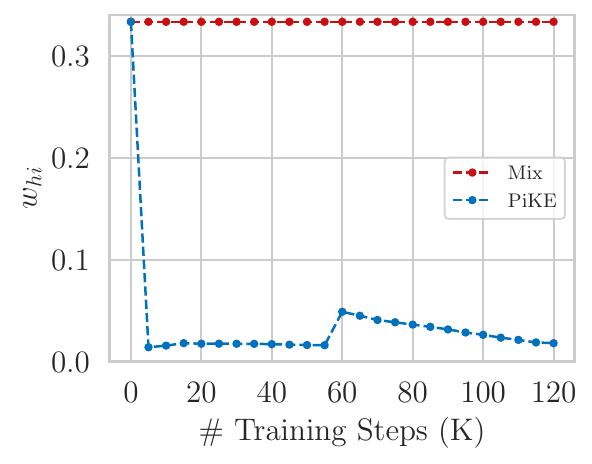}
\end{minipage}
\caption{The sampling weights for each dataset during the pre-training of 1B GPT-2-style multilingual language models on mC4 (English), mC4 (Hindi), and mC4 (German). Here, $w_{\text{en}}$ represents the sampling weight for the English dataset, $w_{\text{hi}}$ for the Hindi dataset, and $w_{\text{de}}$ for the German dataset.}
\label{fig:pike_sampling_weights}
\end{figure*}

Figure~\ref{fig:pike_sampling_weights} illustrates how the adaptive sampling weights of PiKE evolve during language model pre-training. Compared to the Mix sampling strategy, which assigns equal sampling weights to each task, PiKE adaptively adjusts the sampling weights among English, German, and Hindi by leveraging the positive interaction of task gradients. This adaptive data selection allows PiKE to achieve superior performance compared to fixed or heuristic-based baselines.

\section{Derivations and Proofs} \label{app: theory}

\subsection{Detailed Derivation of~\eqref{eq:ExampleExpectedLoss}} \label{sec:derivationExampleExpectedLoss}
Recall that 
\[
\bg_t = \frac{1}{b_1+b_2}\left(b_1e_1e_1^\top + b_2 e_2 e_2^\top\right)\ttheta_t + \bz,
\]
Then
\begin{align}
    \ttheta_{t+1} &= \ttheta_{t} - \eta \frac{1}{b_1+b_2}\left(b_1e_1e_1^\top + b_2 e_2 e_2^\top\right)\ttheta_t - \eta \bz \nonumber \\ 
    & = \ttheta_t - \frac{\eta}{b} 
        \begin{bmatrix}
        b_1 & 0 \\
        0 & b_2 \\
        \end{bmatrix} \ttheta_t - \eta\bz \nonumber
\end{align}
Now consider the loss functions for task 1, $\cL_1(\theta_{t+1})$, and task 2, $\cL_2(\theta_{t+1})$, separately, taking the expectation over the randomness of $\mathbf{z}$
{\allowdisplaybreaks
\begin{align}
    \E[\cL_1(\ttheta_{t+1})]) &= \E\left[\frac{1}{2}(\e_1^\top\ttheta_{t+1})^2\right] \nonumber \\
    &= \E\left[\frac{1}{2}\left(\e_1^\top\begin{bmatrix}
        1-\frac{\eta b_1}{b}& 0 \\
        0 & 1-\frac{\eta b_2}{b} \\
        \end{bmatrix}\ttheta_t - \e_1^\top\eta\bz \right)^2\right] \nonumber \\
    &= \frac{1}{2} \left(\begin{bmatrix}
        1-\frac{\eta b_1}{b} & 0 
    \end{bmatrix} \ttheta^\top\right)^2 + \frac{1}{2}\eta^2\e_1^\top\bQ\e_1 \nonumber \\
    &= \frac{1}{2} \left( \left(1-\frac{\eta b_1}{b}\right) \theta_{1,t}\right)^2 + \frac{1}{2}\eta^2\e_1^\top\bQ\e_1 \nonumber 
\end{align}
}
Similarly, for task 2, we have
\begin{align}
    \E[\cL_2(\ttheta_{t+1})]) &= \frac{1}{2} \left( \left(1-\frac{\eta b_2}{b}\right) \theta_{2,t}\right)^2 + \frac{1}{2}\eta^2\e_2^\top\bQ\e_2 \nonumber 
\end{align}
where $\theta_{1,t}$ and $\theta_{2,t}$ denote the first and second component of the vector $\ttheta_t$. Combining the losses for both tasks, the total expected loss becomes
\begin{align}
    \E[\cL(\ttheta_{t+1})] &= \E[\cL_1(\ttheta_{t+1})]) + \E[\cL_2(\ttheta_{t+1})]) \nonumber \\
    &= \frac{1}{2} \left( \left(1-\frac{\eta b_1}{b}\right) \theta_{1,t}\right)^2 + \frac{1}{2} \left( \left(1-\frac{\eta b_2}{b}\right) \theta_{2,t}\right)^2 + \eta^2\frac{b_1\sigma_1^2+b_2\sigma_2^2}{b^2} \nonumber \\
    &= \frac{1}{2}(1- \frac{\eta b_1}{b})^2\theta_{1,t}^2 +\frac{1}{2}(1-\frac{\eta b_2}{b})^2\theta_{2,t}^2 +\eta^2\frac{b_1\sigma_1^2+b_2\sigma_2^2}{b^2}, \nonumber
\end{align}
which completes the derivations.

\subsection{PiKE: Main Theoretical Results}

\vspace{0.3cm}
\begin{lemma}\label{le: correlation_of_losses}
   Assume $\frac{1}{2(K-1)} > \ulc$. If $\|\nabla \cL(\ttheta)\|^2 \leq \epsilon$, we have 
   \[
           \sum_{k=1}^K \|\nabla \cL_k(\ttheta)\|^2 \leq \frac{\epsilon}{1 - 2\,\ulc\,(K-1)}.
   \]
Conversely, if $\|\nabla \cL_k(\ttheta)\|^2 \leq \delta_k,\;\forall k$, then 
\[
        \|\nabla \cL(\ttheta)\|^2 \leq (1-\barc)\sum_{k=1}^K \delta_k \;+\; \barc \left( \sum_{k=1}^K \sqrt{\delta_k} \right)^2 
\]
\end{lemma}
{\it Proof:} 
We first prove the first direction. Notice that 
\begin{align}
    \|\nabla \cL(\ttheta)\|^2 &= \|\sum^K_{k=1} \nabla\cL_k(\ttheta) \|^2 \nonumber \\
    & = \sum^K_{k=1} \|\nabla \cL_k(\ttheta)\|^2 + \sum^K_{k=1}\sum_{j\neq k} \lin{\nabla\cL_j(\ttheta), \nabla \cL_k(\ttheta)} \leq \epsilon \nonumber
\end{align}
where we use the definition of $\nabla\cL(\ttheta)$ and expand the term. Then we have
\begin{align}
    \sum^K_{k=1} \|\nabla \cL_k(\ttheta)\|^2 + \sum^K_{k=1}\sum_{j\neq k} \lin{\nabla\cL_j(\ttheta), \nabla \cL_k(\ttheta)} &\stackrel{(a)}{\geq} \sum^K_{k=1} \|\nabla\cL_k(\ttheta)\|^2 - \ulc \sum^K_{k=1}\sum_{j\neq k}\left( \|\nabla\cL_j(\ttheta)\|^2 + \|\nabla\cL_k(\ttheta)\|^2\right) \nonumber \\
    &\stackrel{(b)}{\geq} \sum^K_{k=1} \|\nabla\cL_k(\ttheta)\|^2\left(1-2\ulc(K-1)\right) \nonumber
\end{align}
where $(a)$ uses the Definition~\ref{def:Interaction_LB}, $(b)$ uses symmetric identity. Thus we get
\[
  \sum_{k=1}^K \|\nabla \cL_k(\ttheta)\|^2 \leq \frac{\epsilon}{1 - 2\,\ulc\,(K-1)}
\]
This completes the proof of the first inequality. We now prove the second inequality. Notice that
\begin{align}
    \|\nabla \cL(\ttheta)\|^2=\|\sum^K_{k=1} \nabla \cL_k(\ttheta)\|^2 &= \sum^K_{k=1} \|\nabla \cL_k(\ttheta)\|^2 + \sum^K_{k=1} \sum_{j\neq k} \lin{\nabla\cL_j(\ttheta), \nabla\cL_k(\ttheta)} \nonumber \\
    & \stackrel{(a)}{\leq} \|\nabla\cL_k(\ttheta)\|^2 + \barc \sum^K_{k=1} \sum_{j\neq k}\|\nabla\cL_j(\ttheta)\|^2 \|\cL_k(\ttheta)\|^2 \nonumber \\
    & = (1-\barc) \|\nabla\cL_k(\ttheta)\|^2 + \barc \|\nabla\cL_k\|^2 + \barc \sum^K_{k=1} \sum_{j\neq k}\|\nabla\cL_j(\ttheta)\|^2 \|\cL_k(\ttheta)\|^2 \nonumber \\
    & \stackrel{(b)}{\leq} (1-\barc) \sum^K_{k=1} \delta_k + \barc\left(\sum^K_{k=1}\sqrt{\delta_k}\right)^2 \nonumber
\end{align}
where $(a)$ use the Definition~\ref{def:Interaction_UB} and $(b)$ combines the second and third terms and use the condition that $\|\nabla \cL_k(\ttheta)\|^2 \leq \delta_k$. This completes the proof of the second inequality.

\vspace{0.3cm}
\begin{lemma}\label{le:optimal_w_for_relax_problem}
    For the  optimization problem 
    \begin{equation} 
    \begin{split}
        \min_{w_1,\ldots,w_K} \quad &\sum^K_{k=1}  w_k \lambda_k + \frac{1}{2} w_k^2\kappa_k\\\
        \textrm{s.t.} \quad &\sum_{k=1}^Kw_k = 1, \quad w_k \geq 0,\quad  \forall k
    \end{split}
    \end{equation}
    the optimal solution is \begin{equation}
    w_k^* = \max\left\{0, -\frac{\mu + \lambda_k}{\kappa_k}\right\} 
    \end{equation}
    where $\mu$ is chosen such that $\sum_{k=1}^K w_k^* = 1$ 
\end{lemma}
{\it Proof:}
Consider the Lagrangian function
\[
    \cL(w_1,\ldots,w_k,\mu,\alpha_1,\ldots,\alpha_k)=\sum^K_{k=1}  w_k \lambda_k + \frac{1}{2} w_k^2\kappa_k + \mu \left(\sum^K_{k=1}w_k - 1\right) -\sum^K_{k=1}\alpha_kw_k
\]
where $\mu$ is Lagrange multiplier for the equality constraint for the constraint $\sum^K_{k=1}w_k=1$ and $\alpha_k\geq0$ are Lagrange multipliers for the nonnegativity constraints $w_k$. Take the partial derivative of $\mathcal{L}$ with respect to $w_k$ and set it to 0:
\[
\frac{\partial \mathcal{L}}{\partial w_k}=\lambda_k+w_k \kappa_k+\mu-\alpha_k = 0
\]
From the Karush-Kuhn-Tucker (KKT) conditions, we also have $w_k^\star \geq 0, \alpha_k \geq 0$, and $\alpha_k w_k^\star=0$. 
If $w_k^\star>0$, then $\alpha_k=0$, which implies
\[
0=\lambda_k+w_k^\star \kappa_k+\mu \quad \Longrightarrow \quad w_k^\star=-\frac{\mu+\lambda_k}{\kappa_k}
\]

If $-\left(\mu+\lambda_k\right) / \kappa_k$ is negative, then $w_k^\star=0$ must hold. Combining these, we get
\[
w_k^*=\max \left\{0,-\frac{\mu+\lambda_k}{\kappa_k}\right\}
\]
Finally, the Lagrange multiplier $\mu$ is determined by enforcing the equality constraint:
\[
\sum_{k=1}^K w_k^*=1
\]
with $\mu$ chosen so that the $w_k^*$ sum to 1 . This completes the proof.

\vspace{0.3cm}

\begin{theorem} 
\label{thm: formal_descent_lemma}(Formal Statement of Theorem~\ref{thm:DescentMainBody})
Suppose Assumption~\ref{as:assumption1} is satisfied. Assume that at the given point~$\ttheta_t$ the gradients are $\ulc$-conflicted and $\barc$-aligned with
$\ulc <\frac{1}{K -2 + b/b_k}, \forall k$. Moreover, assume batching is performed based on the mix strategy~\eqref{eq: mix_framework}, i.e., 
\[
\mathbf{g}_t = \frac{1}{b} \sum_{k=1}^{K} \sum_{i=1}^{b_k} \nabla \ell_k (\boldsymbol{\theta}_t; x_{k,i}) = \sum_{k=1}^K \frac{b_k}{b} \bar{\mathbf{g}}_{t,k},
\]

Then, we have
    \begin{align} 
        \E[\cL(\ttheta_t - \eta \g_t)\,| \, \mathcal{F}_t] \leq \cL(\ttheta_t) + \sum^K_{k=1} b_k\Big(-\frac{\eta}{b}\beta\|\nabla\cL_k(\ttheta_t)\|^2  + \frac{L\eta^2}{2b^2}\sigma_k^2\Big) + \sum^K_{k=1} b_k^2 \frac{L\eta^2}{2b^2}\gamma \|\nabla \cL_k(\ttheta_t)\|^2 \nonumber 
    \end{align}
    where $0\leq \beta \triangleq \min_{k} (1+\ulc(-K+2-\frac{b}{b_k}))$ and $\gamma \triangleq 1 + \barc (K-1)$,  the expectation is taken over batch sampling randomness under the mix strategy $(b_1, \ldots, b_K)$, and $\,| \, \mathcal{F}_t$ denotes the natural filtration of our process.
\end{theorem}

{\it Proof:} 
We begin by revisiting the multi-task optimization problem under consideration. The objective is defined as:
\begin{equation}
\min_{\ttheta \in \R^d} \;\; \cL(\ttheta) := \sum_{k=1}^K \E_{x \sim \cD_k} \left[\ell_k(\ttheta; x)\right],
\end{equation}
where \(\cL(\ttheta)\) is the  aggregate loss over all tasks. Assume we mix the gradients with taking $b_k$ i.i.d. samples from task $k$ for $k=1,\ldots, K$. Then under, Assumption~\ref{as:assumption1} and based on~\eqref{eq: mix_framework}, the estimated gradient direction is given by

\begin{align}
\bg_t &= \frac{1}{\sum_{k=1}^K b_k} \left(\sum_{k=1}^K \sum_{\stackrel{i=1}{x_i\sim \cD_k}}^{b_k} \nabla \ell_k(\ttheta_t; x_i) \right) 
\end{align}

Let $\ttheta_{t+1}$ be the updated point after gradient descent with $\ttheta_{t+1} = \ttheta_t - \eta \bg_t$. By the descent lemma, the following inequality holds for the updated parameter $\ttheta^{+}$:
\begin{equation}
\cL(\ttheta_{t+1}) \leq \cL(\ttheta_t) - \eta \bg_t^\top \nabla \cL(\ttheta_t) + \frac{L \eta^2}{2} \|\bg_t\|^2,
\end{equation}

Conditioned on the natural filtration process ~$\mathcal{F}_t$, we take the expectation over the randomness of the samples draw and obtain:
\begin{align}
\mathbb{E}\left[\cL(\ttheta_{t+1}) \,| \, \mathcal{F}_t \right] 
&\leq \cL(\ttheta_t) -\eta \E[\bg_t\,| \, \mathcal{F}_t]^\top \nabla \cL(\ttheta_t) + \frac{L\eta^2}{2} \E\left(\|\bg_t\|^2\,| \, \mathcal{F}_t\right)\nonumber\\
&\stackrel{(a)}{=} \cL(\ttheta_t) -\eta \left(\frac{1}{b} \sum_{k=1}^K b_k \nabla \cL_k(\ttheta_t)\right)^\top \left( \sum_{k=1}^K\nabla \cL_k(\ttheta_t) \right)\nonumber\\
& \quad + \frac{L\eta^2}{2b^2}
\left(
 \left\|\sum_{k=1}^K b_k \nabla \cL_k(\ttheta_t)\right\|^2 + \sum_{k=1}^K (b_k \sigma_k^2)
\right)
\nonumber\\
&\stackrel{(b)}{=} \cL(\ttheta_t) -
\frac{\eta}{b} \left( \sum_{k=1}^K b_k \|\nabla \cL_k(\ttheta_t)\|^2 + \sum_{k=1}^K\sum_{j\neq k}b_k \lin{\nabla \cL_j(\ttheta_t) ,\nabla \cL_k(\ttheta_t)}\right)\nonumber\\
& \quad + \frac{L\eta^2}{2b^2}
 \left(\left\|\sum_{k=1}^K b_k \nabla \cL_k(\ttheta_t)\right\|^2 + \sum_{k=1}^K (b_k \sigma_k^2)
\right),
\nonumber
\end{align}

where \((a)\) substitutes the definition of \(\bg_t\) and uses the Assumption~\ref{as:assumption1}, and \((b)\) expands the terms. The notation $ \, | \,\mathcal{F}_t$ denotes the natural filtration of the random process (loosely speaking, conditioned on the current time~$t$ and the past history). Continuing our simplifications,  we have 
{\allowdisplaybreaks
\begin{align}
    \mathbb{E}\left[\cL(\ttheta_{t+1}) \,| \, \mathcal{F}_t\right] 
    &\stackrel{(a)}{\leq}   \cL(\ttheta_t) -
    \frac{\eta}{b} \left( \sum_{k=1}^K b_k \|\nabla \cL_k(\ttheta_t)\|^2 - \sum_{k=1}^K\sum_{j\neq k}b_k \ulc  (\|\nabla \cL_j(\ttheta_t)\|^2 + \|\nabla \cL_{k}(\ttheta_t)\|^2)\right)\nonumber\\
    & \quad + \frac{L\eta^2}{2b^2}
    \left(
        \sum^K_{k=1}b_k^2\|\nabla \cL_k(\ttheta_t)\|^2 + \sum^K_{k=1}\sum_{j\neq k}b_jb_k\lin{\nabla\cL_j(\ttheta_t), \nabla\cL_k(\ttheta_t)} + \sum^K_{k=1} b_k\sigma_k^2
    \right), \nonumber\\
    & \stackrel{(b)}{=}  \cL(\ttheta_t) - \frac{\eta}{b} \left( \sum^K_{k=1} \left(b_k-\ulc b_k(K-1) - \ulc \sum_{j\neq k} b_j \right) \|\nabla\cL_k(\ttheta_t) \|^2 \right) \nonumber \\
    & \quad + \frac{L\eta^2}{2b^2}
    \left(
    \sum^K_{k=1}b_k^2\|\nabla \cL_k(\ttheta)\|^2 + \sum^K_{k=1}\sum_{j\neq k}b_jb_k\lin{\nabla\cL_j(\ttheta_t), \nabla\cL_k(\ttheta_t)} + \sum^K_{k=1} b_k\sigma_k^2
    \right), \nonumber\\
    & \stackrel{(c)}{\leq} \cL(\ttheta_t) -  \frac{\eta}{b} \left( \sum^K_{k=1} (b_k - \ulc (K-1) b_k - \ulc(b-b_k)) \|\nabla\cL_k(\ttheta_t) \|^2 \right) \nonumber \\
    & \quad + \frac{L\eta^2}{2b^2}
    \left(
        \sum^K_{k=1}b_k^2\|\nabla \cL_k(\ttheta_t)\|^2 + \sum^K_{k=1}\sum_{j\neq k}\barc b_jb_k\|\nabla\cL_j(\ttheta_t)\|^2 \|\nabla\cL_j(\ttheta_t) \|^2 + \sum^K_{k=1} b_k\sigma_k^2
    \right) \nonumber \\
    & \stackrel{(d)}{=} \cL(\ttheta_t) -  \frac{\eta}{b} \left( \sum^K_{k=1} b_k(1+\ulc(-K+2- b/b_k)) \|\nabla\cL_k(\ttheta_t) \|^2 \right) \nonumber \\
    & \quad + \frac{L\eta^2}{2b^2}
    \left(
        \barc\left(\sum^K_{k=1}b_k\|\nabla\cL_k(\ttheta_t)\|\right)^2 + (1-\barc) \sum^K_{k=1} b_k^2\|\nabla\cL_k(\ttheta_t)\|^2 + \sum^K_{k=1}b_k\sigma_k^2
    \right) \nonumber \\
    & \stackrel{(e)}{\leq} \cL(\ttheta_t) -  \frac{\eta}{b} \left( \sum^K_{k=1} b_k(1+\ulc(-K+2-b/b_k)) \|\nabla\cL_k(\ttheta_t) \|^2 \right) \nonumber \\
    & \quad + \frac{L\eta^2}{2b^2}
    \left(
        \barc K\sum^K_{k=1} b_k^2 \|\nabla \cL_k(\ttheta_t)\|^2 + (1-\barc) \sum^K_{k=1} b_k^2\|\nabla\cL_k(\ttheta_t)\|^2 + \sum^K_{k=1}b_k\sigma_k^2
    \right) \nonumber \\
    & = \cL(\ttheta_t) -  \frac{\eta}{b} \left( \sum^K_{k=1} b_k(1+\ulc(-K+2-b/b_k)) \|\nabla\cL_k(\ttheta_t) \|^2 \right) \nonumber \\
    & \quad + \frac{L\eta^2}{2b^2}
    \left(
        (1-\barc + \barc K) \sum^K_{k=1} b_k^2\|\nabla \cL_k(\ttheta_t)\|^2 + \sum^K_{k=1}b_k\sigma_k^2
    \right)
\end{align}
}
where $(a)$ applies Definition~\ref{def:Interaction_LB} to the second term and expands the third term, $(b)$ expands the summation in the second term, $(c)$ uses the identity $\sum_{k=1}^K \sum_{j \neq k} b_j = \sum_{k=1}^K (b - b_k)$ in the second term and applies Definition~\ref{def:Interaction_UB} to the third term, $(d)$ combines terms in the third term, and $(e)$ uses the inequality $\|\sum_{i=1}^N u_i\|^2 \leq N\sum_{i=1}^N u_i^2$, where $\bu$ is a column vector. We define $\beta$ and $\gamma$ such that
\begin{align}
    \beta &= \min_{k} (1+\ulc(-K+2-\frac{b}{b_k})) \nonumber \\
    \gamma &= 1 + \barc (K-1) \nonumber
\end{align}
Then using the definition of $\beta$ and $\gamma$, substituting back we have
\begin{align}
    \mathbb{E}\left[\cL(\ttheta_{t+1}) \,| \, \mathcal{F}_t\right] 
    & \leq \cL(\ttheta_t) -  \frac{\eta\beta}{b} \left( \sum^K_{k=1} b_k
    \|\nabla\cL_k(\ttheta_t) \|^2 \right) + \frac{L\eta^2}{2b^2}
    \left(
       \gamma \sum^K_{k=1} b_k^2\|\nabla \cL_k(\ttheta_t)\|^2 + \sum^K_{k=1}b_k\sigma_k^2
    \right) \nonumber \\
    & = \cL(\ttheta_t) + \sum^K_{k=1} b_k \left( -\frac{\eta \beta}{b}\|\nabla\cL_k(\ttheta_t)\|^2 + \frac{L\eta^2}{2b^2}\sigma_k^2\right) +\sum^K_{k=1} b_k^2 \frac{L\eta^2}{2b^2}\gamma\|\nabla\cL_k(\ttheta_t)\|^2, \nonumber 
\end{align}
which  completes the proof.

\begin{theorem}\label{thm:DescentLowerBound}
There exist loss functions $\{\ell_k(\cdot,\cdot)\}_{k=1}^K$ satisfying Assumption~\ref{as:assumption1}, whose gradients are $\ulc$-conflicted and $\barc$-aligned with $\ulc = \barc = 0$. For these losses, the upper bound in~\eqref{eq: theorem1} is tight when gradients are computed using the mix strategy~\eqref{eq: mix_framework}; that is, the inequality in~\eqref{eq: theorem1} holds with equality.
\end{theorem}
\begin{proof}The proof is by simply generalizing Example~\ref{example:Adaptive} and showing the steps in the proof of Theorem~\ref{thm: formal_descent_lemma} are tight for this example. Consider the multi-task learning problem where the loss of task~$k$ is given by
\[
\ell_k(\ttheta,x_k ) = \frac{L}{2} (\mathbf{e}_k^\top \ttheta)^2 + \bx_k^\top \ttheta,
\]
with $\bx_k \sim \mathcal{N} (\mathbf{0}, \frac{\sigma_k^2}{d} \mathbf{I})$ is the data for task~$k$. It is easy to verify that this loss is smooth with smoothness parameter $L$. Moreover, we can show that Assumption~\ref{as:assumption1} is satisfied. Notice that
\[
\cL_k(\ttheta) = \mathbb{E} \left[ \ell_k(\ttheta,x_k)\right] = \frac{L}{2} (\mathbf{e}_k^\top \ttheta)^2
\]
and 
\[
\cL(\ttheta) = \sum_{k=1}^K \cL_k(\ttheta) = \frac{L}{2} \|\ttheta\|^2.
\]
First notice that it is easy to check that  $\ulc=\barc = 0$. Let $\gg_t$ be the direction obtained by~\eqref{eq: mix_framework} and assume $\ttheta_{t+1} = \ttheta_t - \eta \gg_t$. Using the form of $\cL(\ttheta)$, one can easily show that (by expanding the $\ell_2$-loss):
\begin{equation}
\cL(\ttheta_{t+1}) =  \cL(\ttheta_t) - \eta \bg_t^\top \nabla \cL(\ttheta_t) + \frac{L \eta^2}{2} \|\bg_t\|^2.
\end{equation}
Conditioned on the natural filtration process~$\mathcal{F}_t$, we  obtain:
\begin{align}
\mathbb{E}\left[\cL(\ttheta_{t+1}) \,| \, \mathcal{F}_t \right] 
&= \cL(\ttheta_t) -\eta \E[\bg_t\,| \, \mathcal{F}_t]^\top \nabla \cL(\ttheta_t) + \frac{L\eta^2}{2} \E\left(\|\bg_t\|^2\,| \, \mathcal{F}_t\right)\nonumber\\
&\stackrel{(a)}{=} \cL(\ttheta_t) -\eta \left(\frac{1}{b} \sum_{k=1}^K b_k \nabla \cL_k(\ttheta_t)\right)^\top \left( \sum_{k=1}^K\nabla \cL_k(\ttheta_t) \right)\nonumber\\
& \quad + \frac{L\eta^2}{2b^2}
\left(
 \left\|\sum_{k=1}^K b_k \nabla \cL_k(\ttheta_t)\right\|^2 + \sum_{k=1}^K (b_k \sigma_k^2)
\right)
\nonumber\\
&\stackrel{(b)}{=} \cL(\ttheta_t) -
\frac{\eta}{b} \left( \sum_{k=1}^K b_k \|\nabla \cL_k(\ttheta_t)\|^2 \right) \nonumber \\
& \quad + \frac{L\eta^2}{2b^2}
 \left(\sum_{k=1}^K \left\| b_k \nabla \cL_k(\ttheta_t)\right\|^2 + \sum_{k=1}^K (b_k \sigma_k^2)
\right),
\nonumber
\end{align}
where \((a)\) substitutes the definition of \(\bg_t\) and uses the Assumption~\ref{as:assumption1}, and \((b)\) is because $ \lin{\nabla\cL_j(\ttheta_t), \nabla\cL_k(\ttheta_t)}= 0, \;\forall j\neq k $. Since $\beta = \gamma = 1$ in this example, we can write
\begin{equation} \nonumber
    \mathbb{E}\left[\cL(\ttheta_{t+1}) \,| \, \mathcal{F}_t\right] 
 = \cL(\ttheta_t) + \sum^K_{k=1} b_k \left( -\frac{\eta \beta}{b}\|\nabla\cL_k(\ttheta_t)\|^2 + \frac{L\eta^2}{2b^2}\sigma_k^2\right) +\sum^K_{k=1} b_k^2 \frac{L\eta^2}{2b^2}\gamma\|\nabla\cL_k(\ttheta_t)\|^2,
\end{equation}
which shows our upper-bound is tight in this example and  completes the proof.
\end{proof}

\begin{theorem}%
\label{thm:app_IterationComplexityConceptualPiKe} (Theorem~\ref{thm:IterationComplexityConceptualPiKe} in the main body)
Suppose the assumptions in Theorem~\ref{thm: formal_descent_lemma} is satisfied and we run the Conceptual PiKE Algorithm (Algorithm~\ref{alg: Basic PiKE}) initialized at $\ttheta_0$ with the SGD optimizer in Step 10 of the algorithm. Let $\Delta_L = \cL(\ttheta_0) - \min_{\ttheta}\cL(\ttheta)$ and $\sigma_{\max} = \max_k \sigma_k$. Suppose $\delta>0$ is a given constant and the stepsize $\eta \leq \frac{\beta \delta}{L\sigma_{\max}^2/b + L\eta \delta}$. Then, after $T = \frac{2\beta \Delta_L}{\eta \delta}$ iterations, Algorithm~\ref{alg: Basic PiKE} finds a point $\bar{\ttheta}$ such that 
\begin{equation}
    \label{eq:app_boundedNormGrad}
    \mathbb{E}\|\nabla \cL_k(\bar{\ttheta})\|^2 \leq \delta,\quad \forall k=1,\ldots, K.
\end{equation}
Moreover, if we choose $\eta = \frac{\beta \delta}{L\sigma_{\max}^2/b + L\eta \delta}$, then the Conceptual PiKE algorithm requires at most 
\[
\bar{T} = \frac{2L\Delta_L(\sigma_{\max}^2/b + \gamma \delta)}{\delta^2 \beta^2}
\]
iterations to find a point satisfying~\eqref{eq:app_boundedNormGrad}.
\end{theorem}

{\it Proof:}
We prove this by contradiction. Assume that $\max_k \mathbb{E}\|\nabla\cL_k(\ttheta_t)\|^2 > \delta$ for $t=0,\ldots,T-1$. First notice that Theorem~\ref{thm: formal_descent_lemma} implies that for all t, we have 
\begin{align}\label{eq:app_Iteration_eq_1}
     \E[\cL(\ttheta_{t+1}) \, | \, \mathcal{F}_t] & \leq \cL(\ttheta_t) + \sum^K_{k=1}w_k^\star\left(-\eta\beta\|\nabla\cL_k(\ttheta_t))\|^2 + \frac{L\eta^2\sigma^2_{\text{max}}}{2b}\right)  \nonumber \\ 
     & \quad \quad \quad \quad + \sum^K_{k=1}\frac{w_k^\star}{2}\left(L\eta^2\gamma\|\nabla\cL_k(\ttheta_t)\|^2\right)
\end{align}
where $\{w_k^\star\}^K_{k=1}$ is the minimizer of the RHS of the \eqref{eq:app_Iteration_eq_1} on the constrained set $\{(w_1,\ldots,w_k) | \sum^K_{k=1}w_k=1,\;w_k\geq0\;\forall k\in K\}$. Since $w_k^\star$ is the minimizer of the RHS of \eqref{eq:app_Iteration_eq_1}, we have 
\begin{align}\label{eq:app_Iteration_eq_2}
    w_k^\star\left(-\eta\beta\|\nabla\cL_k(\ttheta_t)\|^2 + \frac{L\eta^2}{2b}\sigma_{\text{max}}^2\right) & + \frac{w_k^\star}{2}L\eta^2\gamma\|\nabla\cL_k(\ttheta_t)\|^2  \nonumber \\ 
    & \leq \left(-\eta\beta\|\nabla\cL_{k_t^\star}(\ttheta_t)\|^2 + \frac{L\eta^2}{2b}\sigma_{\text{max}}^2\right) + \frac{L\eta^2}{2}\gamma\|\nabla\cL_{k_t^\star}(\ttheta_t)\|^2
\end{align}

where $k^\star_t \in \arg\max_k\|\nabla\cL_k(\ttheta_t)\|^2$. Moreover, we have
\[
    \eta \leq \frac{\beta \delta}{L\frac{\sigma_{\text{max}}^2}{b}+L\gamma \delta}\leq  \frac{\beta \|\nabla\cL_{k^\star_t }(\ttheta)\|^2}{L\frac{\sigma_{\text{max}}^2}{b}+L\gamma\|\nabla\cL_{k^\star_t }(\ttheta_t)\|^2},
\] 
Therefore,  
\begin{equation}\label{eq:app_Iteration_eq_3}
    \left(-\eta\beta\|\nabla\cL_{k^\star_t}(\ttheta_t)\|^2 + \frac{L\eta^2}{2b}\sigma_{\text{max}}^2\right) + \frac{L\eta^2}{2}\gamma\|\nabla\cL_{k_t^\star}(\ttheta_t)\|^2 \leq -\frac{\beta\eta}{2}\|\nabla\cL_{k^\star_t}(\ttheta_t)\|^2
\end{equation}
Combining equation (\ref{eq:app_Iteration_eq_1}), (\ref{eq:app_Iteration_eq_2}), and (\ref{eq:app_Iteration_eq_3}), we obtain
\[
    \mathbb{E}[\cL(\ttheta_{t+1}) \, | \, \mathcal{F}_t] \leq \cL(\ttheta_t) - \frac{\beta \eta}{2} \|\nabla\cL_{k^\star_t}(\ttheta_t)\|^2
\]
Or equivalently,
\[
    \mathbb{E}[\cL(\ttheta_{t+1}) \, | \, \mathcal{F}_t] \leq \cL(\ttheta_t) - \frac{\beta\eta}{2} \max_k\|\nabla\cL_k(\ttheta_t)\|^2
\]
Taking the expectation over all sources of randomness in the algorithm, summing the above inequality from $t = 0$ to $t = T - 1$, and simplifying the resulting telescoping sum, we obtain:
\[
    \mathbb{E} [\cL(\ttheta_T)] \leq \cL(\ttheta_0) - \frac{\beta\eta}{2}\sum^{T-1}_{t=1}\max_k\mathbb{E}[ \|\nabla\cL_k(\ttheta_t)\|^2]
\]
Recalling the contradiction assumption that $ \delta < \mathbb{E} [\|\nabla\cL_k(\ttheta_t)\|^2] $, we get
\[
    \mathbb{E}[\cL(\ttheta_T)] \leq \cL(\ttheta_0) - \frac{\beta\eta}{2} T\delta
\]
Using the definition $\Delta_L \triangleq \cL(\ttheta_0) - \min_{\ttheta} \cL(\ttheta)$, we get 
\[
    T\leq \frac{2\Delta_L}{\beta\eta\delta}
\]

Finally notice that by setting $\eta=\frac{\beta\delta}{L\frac{\sigma_{\text{max}}^2}{b}+L\gamma\delta}$, we get
\[
    T \leq \Bar{T} = \frac{2\Delta_L}{\beta\eta} = \frac{2L\Delta_L}{\beta\delta^2}\left(\frac{\sigma_{\text{max}}^2}{b}+\gamma\delta\right),
\]
which completes the proof. 

\subsection{Conceptual PiKE versus Static Uniform Mix Batching}
\label{app: PiKEvsUniform}
We now perform a standard analysis of Stochastic Gradient Descent (SGD), adapted to our specific setup where uniform (static) mini-batching is used (i.e., when $b_k = b/K,\;\forall k=1,\ldots, K$). As is common in the analysis of gradient-based algorithms in smooth, nonconvex settings, we begin by quantifying the expected decrease in the objective function at each iteration. This is known as the descent per iteration. Once this is established, we use a telescoping sum argument to derive the iteration complexity of the algorithm, which tells us how many steps are needed to achieve a desired level of accuracy, given a properly chosen learning rate.

Let’s begin by analyzing the descent that occurs in a single iteration. Recall that the objective is defined as:
\begin{equation}
\min_{\ttheta \in \R^d} \;\; \cL(\ttheta) := \sum_{k=1}^K \E_{x \sim \cD_k} \left[\ell_k(\ttheta; x)\right].
\end{equation}
Under Assumption~\ref{as:assumption1}, and using uniform static mini-batch sampling—where each source $k$ contributes equally to the batch with size $b_k = b/K$—the estimated gradient direction at iteration $t$ is given by:
\begin{align}
\bg_t &= \frac{1}{b} \left(\sum_{k=1}^K \sum_{\stackrel{i=1}{x_i\sim \cD_k}}^{b/K} \nabla \ell_k(\ttheta_t; x_i) \right) 
\end{align}
This expression aggregates gradients from all sources using equally sized sub-batches, providing a mini-batch estimate of the full gradient. 
Let $\ttheta_{t+1}$ denote the updated parameter after performing a gradient descent step, given by $\ttheta_{t+1} = \ttheta_t - \eta \mathbf{g}_t$. By the descent lemma, the following inequality holds for the updated parameter:
$\ttheta_{t+1}$:
\begin{equation}
\cL(\ttheta_{t+1}) \leq \cL(\ttheta_t) - \eta \bg_t^\top \nabla \cL(\ttheta_t) + \frac{L \eta^2}{2} \|\bg_t\|^2,
\end{equation}
Conditioned on the natural filtration $\mathcal{F}_t$, we take the expectation with respect to the randomness in the sampled data and obtain:
\begin{align}
\mathbb{E}\left[\cL(\ttheta_{t+1}) \,| \, \mathcal{F}_t \right] 
&\leq \cL(\ttheta_t) -\eta \E[\bg_t\,| \, \mathcal{F}_t]^\top \nabla \cL(\ttheta_t) + \frac{L\eta^2}{2} \E\left(\|\bg_t\|^2\,| \, \mathcal{F}_t\right)\nonumber\\
&= \cL(\ttheta_t) -\eta \left(\frac{1}{K} \sum_{k=1}^K \nabla \cL_k(\ttheta_t)\right)^\top \left( \sum_{k=1}^K\nabla \cL_k(\ttheta_t) \right)\nonumber\\
& \quad + \frac{L\eta^2}{2b^2}
\left(
 \left\|\sum_{k=1}^K \frac{b}{K} \nabla \cL_k(\ttheta_t)\right\|^2 + \frac{b}{K}\sum_{k=1}^K  \sigma_k^2
\right)
\nonumber
\end{align}
Observing the fact that $\cL(\ttheta) = \sum_{k=1}^K \cL_k(\ttheta)$, we can write
\begin{equation}
    \mathbb{E}\left[\cL(\ttheta_{t+1}) \,| \, \mathcal{F}_t \right] 
\leq \cL(\ttheta_t) -\frac{\eta}{K}\left\|\nabla \cL (\theta_t)\right\|^2 + \frac{L\eta^2}{2b^2} \left(\frac{b^2}{K^2} \left\|\nabla \cL (\theta_t)\right\|^2  + \frac{b}{K}\sum_{k=1}^K \sigma_k^2\right)
\end{equation}
Define $\sigma^2 := \sum_{k=1}^K \sigma_k^2$. By summing u Rearranging the terms, we obtain:

\begin{equation}
    \mathbb{E}\left[\cL(\ttheta_{t+1}) \,| \, \mathcal{F}_t \right] 
\leq \cL(\ttheta_t) + \left( - \frac{\eta}{K}  + \frac{L\eta^2}{2K^2}\right) \left\|\nabla \cL (\theta_t)\right\|^2 + \frac{L\eta^2}{2bK}\sigma^2
\end{equation}
Taking the expectation over all sources of randomness in the algorithm, summing the above inequality from $t = 0$ to $t = T - 1$, and simplifying the resulting telescoping sum, we obtain:
\begin{equation}
    \mathbb{E}\left[\cL(\ttheta_T) \right] 
\leq \cL(\ttheta_0) + \left( - \frac{\eta}{K}  + \frac{L\eta^2}{2K^2}\right) \sum_{t=0}^{T-1}\mathbb{E}\left[\left\|\nabla \cL (\theta_t)\right\|^2\right] + \frac{LT\eta^2}{2bK}\sigma^2.
\end{equation}
Let $\Delta_L := \mathcal{L}(\ttheta_0) - \min_{\ttheta} \mathcal{L}(\ttheta)$ denote the initial suboptimality gap. As is standard in iteration complexity analysis, we are interested in determining the number of iterations required to reach a point~$\ttheta_{t_0}$ where $\mathbb{E}\left[\left\|\nabla \mathcal{L}(\ttheta_{t_0})\right\|^2\right] \leq \delta$. Suppose, for the moment, that this condition has not yet been satisfied. Then, we can rewrite the above inequality as
\begin{equation}
    0
\leq 
\Delta_L + \left( - \frac{\eta}{K}  + \frac{L\eta^2}{2K^2}\right)  T\delta + \frac{LT\eta^2}{2bK}\sigma^2.
\end{equation}
Equivalently, under a properly chosen step size (to be defined later), we obtain the following upper bound on the number of iterations
\[
T \leq \frac{\Delta_L}{
\left(  \frac{\eta}{K}  - \frac{L\eta^2}{2K^2}\right)  \delta - \frac{L\eta^2}{2bK}\sigma^2
}.
\]
By optimizing the choice of  $\eta$ to minimize the right-hand side, we derive the following bound on the iteration complexity
\[
T \leq \frac{2 L\Delta_L \left(\delta + \frac{\sigma^2}{b}K\right)}{
\delta^2
}.
\]

\vspace{0.2cm}

In summary, for the uniform mix batching strategy, the standard analysis yields the following upper bound on the iteration complexity:
\begin{equation}\label{eq:UniformIterationComplexity}
\bar{T}_{\textrm{Uniform}} = \frac{2 L\Delta_L \left(\delta + \frac{\sigma^2}{b}K\right)}{
\delta^2}.
\end{equation}

In contrast, when the task gradients are nearly orthogonal—i.e., when the alignment/conflict parameters satisfy $\barc,\ulc\approx 0$ —Theorem~\ref{thm:IterationComplexityConceptualPiKe} gives the following iteration complexity bound for the Conceptual PiKE algorithm:
\begin{equation}\label{eq:PiKEIterationComplexity}
\bar{T}_{\textrm{PiKE}} = \frac{2L\Delta_L\left( \delta+ \sigma_{\max}^2/b \right)}{\delta^2},
\end{equation}
which clearly illustrates the advantage of PiKE for this regime in terms of iteration complexity upper-bound.

\subsection{Balanced-PiKE: Theoretical Developments and Derivations}\label{app:fairness_theory}
Here we prove Lemma~\ref{lemma:MinToMinMax} which is the idea behind the Balanced-PiKE algorithm. Before proving it, let us, for the sake of completeness, re-state the lemma
\begin{lemma} [Restatement of Lemma~\ref{lemma:MinToMinMax}]
   Let $0<\tau \in \mathbb{R}$. Then,  the tilted empirical risk minimization problem
    \[
       \min_{\ttheta}  \frac{1}{\tau} \log \left(\sum^K_{k=1} e^{\tau\cL_k(\ttheta)}\right)
    \]
is equivalent to 
    \begin{align}
    \min_{\ttheta} \quad \max_{\substack{\mathbf{y} \in \mathbb{R}_{+}^K \\ \sum_{k=1}^K y_k=\tau}} \sum^K_{k=1} y_k\cL_k(\ttheta) - \sum^K_{k=1} \frac{y_k}{\tau} \log \left(\frac{y_k}{\tau} \right).
    \end{align}  
    Moreover, for any fixed $\ttheta$, the inner maximization problem is maximized at
    $y_k^\star = \frac{\tau e^{\tau\cL_k(\ttheta)}}{\sum^K_{j=1} e^{\tau\cL_j(\ttheta)}}, \; \forall k.$
\end{lemma}
The above lemma is the direct consequence of applying Lemma~\ref{le:tilted_eqivalence} and Lemma~\ref{le:fairness_optimal_y}.
\vspace{0.3cm}
\begin{lemma}\label{le:fairness_optimal_y}
    For the problem  
    \[
    \max _{\substack{\mathbf{y} \in \mathbb{R}_{+}^K \\ \sum_{k=1}^K y_k=\tau}}\left(\sum_{k=1}^K y_k x_k-\sum_{k=1}^K \frac{y_k}{\tau} \log \left(\frac{y_k}{\tau}\right)\right),
    \] the optimal $\mathbf{y}$ is given by
    \[
        y_k^\star =\frac{\tau e^{\tau x_k-1}}{\sum^K_{k=1} e^{\tau x_k -1}}
    \]
\end{lemma}
{\it Proof:}
We start by forming and maximizing the Lagrangian function
\[
    \max _{\substack{\mathbf{y} \in \mathbb{R}_+^K }}\left(\sum_{k=1}^K y_k x_k-\sum_{k=1}^K \frac{y_k}{\tau} \log \left(\frac{y_k}{\tau}\right) + \mu \left(\sum_{k=1}^K y_k - \tau\right)\right)
\] 
where $\mu$ is a free variable. Taking the partial derivative of the objective with respect to $y_k$ and setting it to zero gives 
\[
        y_k^\star =\alpha e^{\tau x_k },
\]
where the coefficient $\alpha$ is independent of the index~$k$ and should be chosen such that $\sum_k y_k^* = 1$, implying
\[
        y_k^\star =\frac{\tau e^{\tau x_k}}{\sum^K_{j=1} e^{\tau x_j }}.
\]

\vspace{0.2cm}

\begin{lemma}\label{le:tilted_eqivalence}
Let $\bx \in \R^K$ and $\tau > 0$. Then,
    \[
         \log \left(\sum_{k=1}^K e^{\tau x_k}\right)=\max_{\substack{\mathbf{y} \in \mathbb{R}_{+}^K \\ \sum_{k=1}^K y_k=\tau}}\left(\sum_{k=1}^K y_k x_k-\sum_{k=1}^K \frac{y_k}{\tau} \log \left(\frac{y_k}{\tau}\right)\right)
    \] 
\end{lemma}
\begin{proof}
The proof is by simply plugging in the optimal value $y^\star$ obtained by \Cref{le:fairness_optimal_y} in the objective function.
\end{proof}

\end{document}